\documentclass[10pt,twocolumn,letterpaper]{article}

\usepackage[pagenumbers]{iccv} 

\definecolor{iccvblue}{rgb}{0.21,0.49,0.74}
\usepackage[pagebackref,breaklinks,colorlinks,allcolors=iccvblue]{hyperref}
\usepackage[accsupp]{axessibility}
\def\paperID{12973} 
\def\confName{ICCV}
\def\confYear{2025}

\title{DictAS: A Framework for Class-Generalizable Few-Shot Anomaly Segmentation via Dictionary Lookup}

\author{
	Zhen Qu\textsuperscript{\rm 1, 2} \quad Xian Tao\textsuperscript{\rm 1,2,3,4}\footnotemark[2]~~ \quad Xinyi Gong\textsuperscript{\rm 5} \quad ShiChen Qu\textsuperscript{\rm 1, 2} \quad
	Xiaopei Zhang\textsuperscript{\rm 7} \\
	Xingang Wang\textsuperscript{\rm 1, 2} \quad Fei Shen\textsuperscript{\rm 1,2,3,4} \quad Zhengtao Zhang\textsuperscript{\rm 1,2,3} \quad Mukesh Prasad\textsuperscript{\rm 6} \quad Guiguang Ding\textsuperscript{\rm 8} \\
	\textsuperscript{\rm 1}Institute of Automation, Chinese Academy of Sciences \\
	\textsuperscript{\rm 2}School of Artificial Intelligence, University of Chinese Academy of Sciences \quad \textsuperscript{\rm 3}Casivision \\
	\textsuperscript{\rm 4}Longmen Laboratory \quad \textsuperscript{\rm 5}HDU \quad \textsuperscript{\rm 6}UTS \quad \textsuperscript{\rm 7}UCLA  \quad \textsuperscript{\rm 8}Tsinghua University\\
}

\usepackage{colortbl}
\usepackage{multirow}
\usepackage{wrapfig}
\usepackage{tabularx}
\usepackage{makecell}
\usepackage{enumitem}
\usepackage{marvosym}
\usepackage{adjustbox}
\usepackage{algorithm}
\usepackage{algorithmic}
\usepackage{pifont}
\usepackage{float} 
\definecolor{lightgreen}{RGB}{240,255,255}
\definecolor{darkred}{RGB}{180,0,0} 

\definecolor{lightblue1}{RGB}{100, 149, 237} 
\definecolor{lightgreen1}{RGB}{34, 139, 34} 
\definecolor{lightorange1}{RGB}{255, 165, 0} 
\usepackage{listings}
\usepackage{xcolor}

\lstdefinestyle{mystyle}{
	language=Python,
	basicstyle=\ttfamily\footnotesize,
	keywordstyle=\color{blue},
	commentstyle=\color{gray},
	stringstyle=\color{red},
	numbers=left, 
	numberstyle=\tiny\color{gray},
	stepnumber=1,    
	breaklines=true,
	frame=lines, 
	captionpos=b,
}
\newcommand{\dev}[1]{\footnotesize{$\pm$#1}}

\begin{document}
\maketitle

\footnotetext[2]{Corresponding Author: taoxian2013@ia.ac.cn.}
\footnotetext[3]{Project Page: \url{https://github.com/xiaozhen228/DictAS}}

\begin{abstract}
Recent vision-language models (e.g., CLIP) have demonstrated remarkable class-generalizable ability to unseen classes in few-shot anomaly segmentation (FSAS), leveraging supervised prompt learning or fine-tuning on seen classes. However, their cross-category generalization largely depends on prior knowledge of real seen anomaly samples. In this paper, we propose a novel framework, namely DictAS, which enables a unified model to detect visual anomalies in unseen object categories without any retraining on the target data, only employing a few normal reference images as visual prompts. The insight behind DictAS is to transfer dictionary lookup capabilities to the FSAS task for unseen classes via self-supervised learning, instead of merely memorizing the normal and abnormal feature patterns from the training set. Specifically, DictAS mainly consists of three components: (1) \textbf{Dictionary Construction} - to simulate the index and content of a real dictionary using features from normal reference images. (2) \textbf{Dictionary Lookup} - to retrieve queried region features from the dictionary via a sparse lookup strategy. When a query feature cannot be retrieved, it is classified as an anomaly. (3) \textbf{Query Discrimination Regularization}- to enhance anomaly discrimination by making abnormal features harder to retrieve from the dictionary. To achieve this, Contrastive Query Constraint and Text Alignment Constraint are further proposed. Extensive experiments on seven public industrial and medical datasets demonstrate that DictAS consistently outperforms state-of-the-art FSAS methods.
\end{abstract}

\section{Introduction}
\label{sec:intro}
Few-shot (few-normal-shot) anomaly segmentation (FSAS) aims to identify anomalous regions in images given only a limited number of normal samples. This task is particularly important in scenarios where training data is scarce and pixel-level annotations are limited, such as industrial defect detection \cite{quTIM, chenTIM, chenTII} and medical image analysis \cite{medical1, medical2, BrasTS}. 
\par 
\begin{figure*}[t]
	\centering
	\includegraphics[width= 1.9\columnwidth]{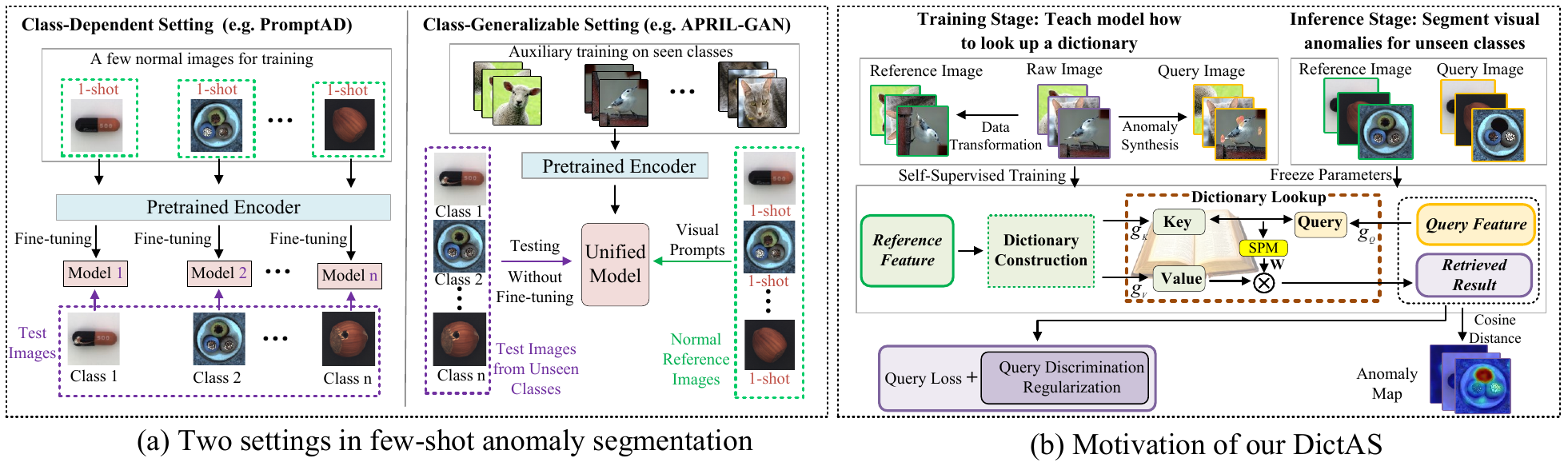}
	\caption{(a) Two settings in few-shot anomaly segmentation. (b) Motivation of our DictAS. }
	\label{Fig1}
\end{figure*}

Existing FSAS methods typically follow class-dependent or class-generalizable setting, as shown in Figure \ref{Fig1}(a). Class-dependent approaches require fine-tuning on each unseen class with a limited number of normal samples. Since the training and testing image classes are identical, they primarily focus on modeling the distribution of normal images or learning the boundary between normal and abnormal content through regularization, without considering the substantial domain gaps across categories \cite{SFAD1, SFAD3, SFAD4, SFAD5, SFAD6, PromptAD}.  Consequently, such methods encounter significant challenges when applied to privacy-sensitive medical applications or dynamic industrial scenarios with frequent production line changes.
\par  
In contrast, as shown in the right part of Figure \ref{Fig1}(a), class-generalizable approaches aim to develop a unified model capable of detecting anomalies in unseen classes without retraining on target data, relying solely on a few normal samples as visual prompts \cite{RegAD, winclip, AnomalyGPT, VAND, fastrecon, MetaUAS}. The earliest work, RegAD \cite{RegAD}, introduces feature registration to align features but suffers from reduced inference efficiency due to heavy reference image augmentation. FastRecon \cite{fastrecon} proposes feature reconstruction using linear regression but is prone to over-reconstruction. More recent methods have increasingly focused on pre-trained vision-language models (VLMs), particularly CLIP \cite{CLIP}, to enhance zero/few-shot generalization for unseen classes \cite{AnomalyCLIP, AdaCLIP, VCP, bayes, winclip, VAND, AnomalyGPT}. Approaches such as WinCLIP \cite{winclip} and APRIL-GAN \cite{VAND} introduce memory bank-based visual priors through normal images and enhance FSAS performance by exploiting CLIP's image-text alignment capabilities. Despite their promising performance, these methods often rely on \textquotedblleft \textit{empirical knowledge}\textquotedblright \ learned from seen abnormal images during auxiliary training stage, which constrains their ability to generalize to novel classes.
 \par  
However, even novice human inspectors can detect anomalies in unseen categories with just a few normal samples as references, without extensive prior experience. We approximate this intuition as a dictionary lookup: if a region in the query image cannot be retrieved from the dictionary, it is classified as anomalous; otherwise, it is normal. Inspired by this, a novel self-supervised framework built on CLIP, namely DictAS, is proposed for class-generalizable FSAS. The framework comprises three components: Dictionary Construction, Dictionary Lookup, and Query Discrimination Regularization, as depicted in Figure \ref{Fig1}(b). Our motivation is to reformulate anomaly segmentation as a dictionary lookup task—determining whether a query feature exists in the dictionary. Through self-supervised training, the model acquires a feature-agnostic and dynamically adaptive dictionary lookup capability, enabling class-generalizable FSAS.
\par 
The proposed Dictionary Construction component organizes normal reference image features into a structured dictionary, where the \textit{Dictionary Key} and \textit{Dictionary Value} serve as the index and content, respectively. As demonstrated in Figure \ref{Fig1}(b), given a query image, its extracted features are transformed into a \textit{Dictionary Query}, which is then matched against the \textit{Dictionary Key} to retrieve the most relevant normal patterns. To refine this retrieval process, we introduce a sparse lookup strategy within the Dictionary Lookup component. Specifically, the \textit{Query-Key} matching results are processed by a Sparse Probability Module (SPM), which encourages sparsity in the retrieval process and prioritizes the most relevant \textit{Dictionary Value}. Unlike prior VLM-based methods \cite{winclip, VAND, AnomalyGPT} that rely on memory-based visual priors or prior knowledge derived from text-image alignment, DictAS learns dynamically adaptive retrieval weights, enabling flexible adaptation to variations in both query and reference images. For self-supervised training, the raw images are processed using anomaly synthesis and data transformation algorithms to generate (query-reference) image pairs. To optimize the dictionary lookup task, a query loss is further introduced to minimize the average distance between the normal regions of input \textit{\textbf{Query Feature}} and their counterparts in \textit{\textbf{Retrieved Result}} generated by the dictionary lookup process.
\par 
To enhance the anomaly discrimination ability, the Query Discrimination Regularization component is proposed, which makes it harder for anomalous regions in the query image to be retrieved from the dictionary. It consists of two parts: the Contrastive Query Constraint (CQC) and the Text Alignment Constraint (TAC). The CQC explicitly enforces greater feature distances between the \textbf{\textit{Query Feature}} and its \textbf{\textit{Retrieved Result}} in anomalous regions compared to normal regions, ensuring that anomalies are effectively distinguished. Meanwhile, the TAC regularizes the retrieved global image representation by aligning it with the normal text embedding space, preventing the model from misinterpreting anomalous content as normal. 
\par 
In summary, our contributions are threefold: 1) We propose DictAS, a novel self-supervised framework that reformulates anomaly segmentation as a dictionary lookup task, enabling models to learn an adaptive retrieval capability for class-generalizable FSAS; 2) We introduce two regularizations strategies—Contrastive Query Constraint and Text Alignment Constraint—to enhance the robustness of dictionary-based querying and improve anomaly discrimination; 3) DictAS achieves state-of-the-art performance on seven industrial and medical datasets, demonstrating superior FSAS performance even when trained on auxiliary datasets without pixel-level annotations.
\section{Related Works}
\label{Related}
 \begin{figure*}[t]
	\centering
	\includegraphics[width= 1.9\columnwidth]{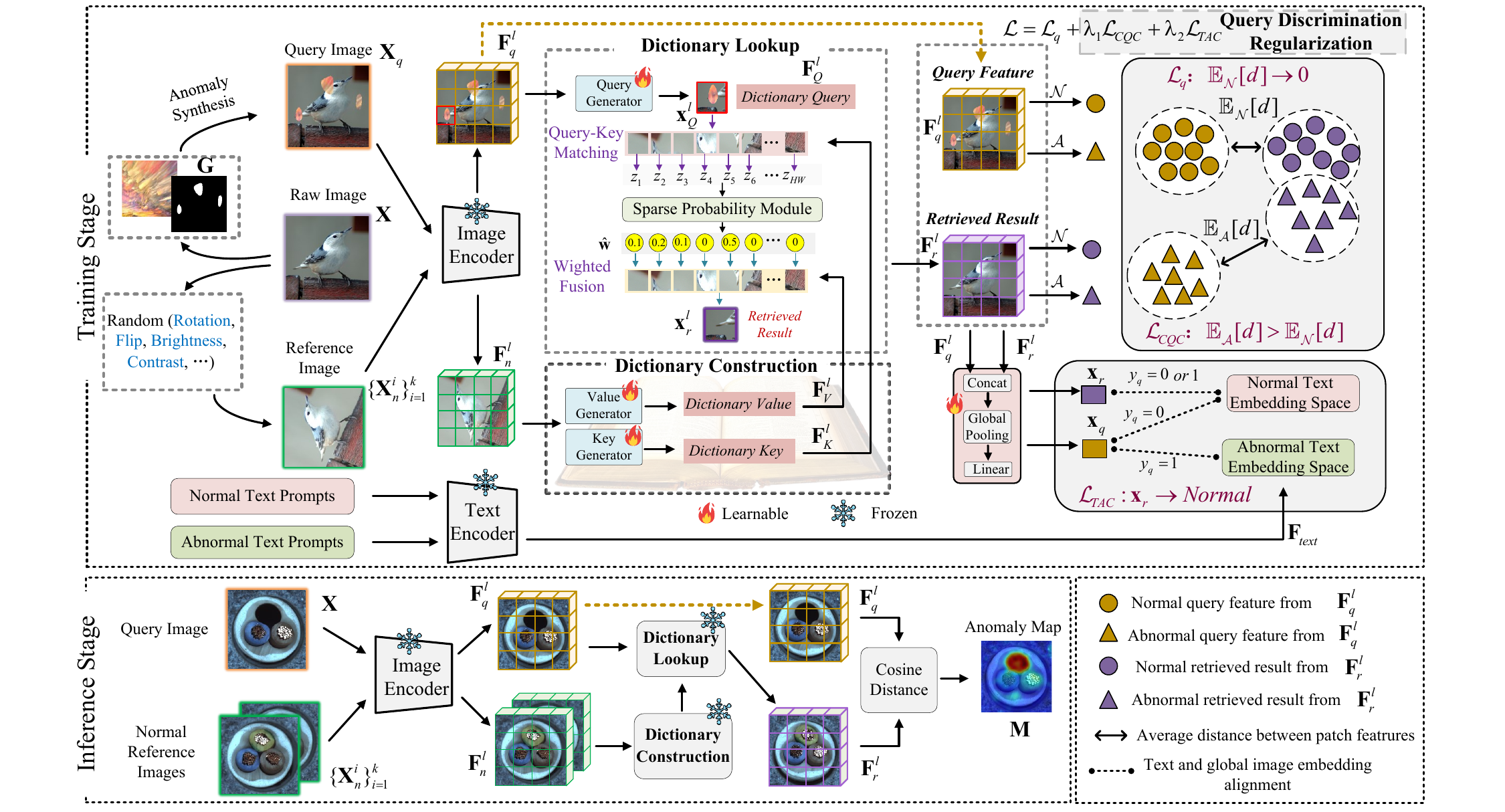}
	\caption{The framework of DictAS. It mainly consists of three components: Dictionary Construction, Dictionary Lookup and Query Discrimination Regularization. During training, the number of reference images is set to $k=1$ for model efficiency. During inference, $k\ge 1$ normal reference images are used as visual prompts.}
	\label{Fig2_Frmawork}
\end{figure*}
\textbf{Class-Dependent FSAS} methods fine-tune a separate model for each class using only its corresponding normal reference images. They primarily focus on: (1) estimating the distribution of a few normal samples \cite{SFAD1, SFAD2}, or (2) learning a discriminative boundary using only normal data \cite{SFAD4, SFAD5, SFAD6, PromptAD}. Since these methods do not need to address domain shifts between training and testing classes, they typically achieve superior performance compared to class-generalizable approaches.
\par
\textbf{Class-Generalizable FSAS} methods train a unified model on seen classes from an auxiliary dataset and directly generalize to unseen classes without additional fine-tuning. Early distance-based approaches \cite{SPADE, padim, patchcore, Graphcore} detect anomalies by measuring the distance between query and reference images or features but are sensitive to image perturbations. Meta-learning-based methods \cite{MetaUAS, RegAD, MetaAno2} aim to achieve class-generalizable FSAS through task generalization but often require offline construction of auxiliary datasets and involve complex training processes. Feature residual-based methods \cite{InctRL, ResAD} mitigate inter-class variations by refining features based on the residuals between query and reference image features, though their effectiveness depends on the quality of residual feature extraction. Vision-language model-based approaches \cite{winclip, VAND} align text embeddings with image patch features in a joint embedding space, enabling language-guided anomaly segmentation. Building upon this, AnomalyGPT \cite{AnomalyGPT} further incorporates large language models to facilitate multi-turn user interaction, but introduces higher inference overhead.
\par
Unlike the aforementioned class-generalizable methods, our DictAS reformulates FSAS as a dictionary lookup task, inspired by human inspectors. This enables us to leverage a large number of pixel-annotation-free images from seen classes for self-supervised training, guiding the model to learn adaptive querying and robust dictionary retrieval, thereby effectively addressing the FSAS task.
\section{Method}

\textbf{Problem Statement.} Class-generalizable FSAS is a challenging task that aims to achieve high performance on unseen classes without requiring fine-tuning on the target data. The seen classes $C^{s}$ in the auxiliary training set and the unseen classes $C^{u}$ in the test set must satisfy $C^s \cap C^u = \emptyset$. The ultimate goal is to develop a unified model capable of segmenting visual anomalies in novel classes $C^{u}$, relying solely on $k$-shot normal samples corresponding to the same class as the visual prompts. 
\par 
\subsection{Overview}
The framework of our DictAS is illustrated in Figure \ref{Fig2_Frmawork}. We employ the transformer-based CLIP model as the backbone, in line with recent FSAS approaches \cite{winclip, VAND, PromptAD}.
\par 
In training, we introduce a novel self-supervised learning strategy that eliminates the need for extensive pixel-labeled training samples. Given a raw image $\mathbf{X} \in \mathbb{R}^{h \times w \times 3}$ from a seen class $C^{s}$, we generate $k$ reference images $\{\mathbf{X}_n^i\}_{i=1}^k$ by applying data transformations (e.g., random rotation, flipping) to simulate normal variations. Simultaneously, a query image $\mathbf{X}_q$ is created by synthesizing anomalies on $\mathbf{X}$ to mimic anomalous scenarios. A  pretrained image encoder extracts multi-layer patch-level features from both $\{\mathbf{X}_n^i\}_{i=1}^k$ and $\mathbf{X}_q$, denoted as $\mathbf{F}_n^l \in \mathbb{R}^{kHW \times C}$ and $\mathbf{F}_q^l \in \mathbb{R}^{HW \times C}$, where $l = 1, 2, \dots, L$, with $H$, $W$, and $C$ denoting the height, width, and feature dimension, respectively. These extracted features are utilized in three key components of our model: 1) \textbf{Dictionary Construction}: The reference image features $\mathbf{F}_n^l$ are used to construct a dictionary for retrieval; 2) \textbf{Dictionary Lookup}: The \textit{\textbf{Query Feature}} $\mathbf{F}_q^l$ serve as queries for retrieval, producing the \textit{\textbf{Retrieved Result}} $\mathbf{F}_r^l$; 3) \textbf{Query Discrimination Regularization}: Two regularization terms are added into the loss function to jointly optimize the model based on $(\mathbf{F}_r^l, \mathbf{F}_q^l)$, improving the discrimination between normal and anomalous patterns.
\par 
During inference, the model constructs a dictionary using features $\mathbf{F}_n^l$ extracted from a few normal reference images of unseen classes. Given a query image $\mathbf{X}_q$ from the same unseen classes, its features $\mathbf{F}_q^l$ are compared with the \textit{\textbf{Retrieved Result}} $\mathbf{F}_r^l$. The computed distance guides the generation of the final anomaly map $\mathbf{M}$.
\subsection{Dictionary Construction}
A well-constructed dictionary typically consists of two components: an index and the corresponding content, referred to as the \textit{Dictionary Key} and \textit{Dictionary Value}, respectively. Motivated by this, we design a Key Generator $g_K$ and a Value Generator $g_V$ to transform the extracted normal reference image features $\mathbf{F}_n^l$ into structured dictionary representations. Meanwhile, the Query Generator $g_Q$ processes the \textit{\textbf{Query Feature}} $\mathbf{F}_q^l$ to obtain a \textit{Dictionary Query}, which is then matched against the \textit{Dictionary Key} for retrieval. The feature transformations are defined as follows: 
	\begin{gather}
	\mathbf{F}_Q^l = g_Q(\mathbf{F}_q^l) = AttnBlock\_Q(\mathbf{F}_q^l)\\
      \mathbf{F}_K^l = g_K(\mathbf{F}_n^l) = AttnBlock\_K(\mathbf{F}_n^l) \\
	  \mathbf{F}_V^l = g_V(\mathbf{F}_n^l) = \mathbf{F}_n^l + AttnBlock\_V(\mathbf{F}_n^l)
\end{gather}
where $\mathbf{F}_Q^l \in\mathbb{R}^{HW\times C}$, $\mathbf{F}_K^l, \mathbf{F}_V^l \in \mathbb{R}^{kHW\times C}$ are \textit{Dictionary Query, Key and Value}, respectively. Here, $AttnBlock$ represents a self-attention-based transformer block designed for feature adaptation. The additional residual connection in $g_V$ helps preserve the fine-grained details of $\mathbf{F}_n^l$, enhancing feature fidelity in the constructed dictionary.
\par 
\textbf{Design of \textit{AttnBlock}.} To enable the dictionary representations to capture meaningful global relationships, we employ a self-attention mechanism within each $AttnBlock$. Given an input feature $\mathbf{F}_{in} \in \{\mathbf{F}_q^l, \mathbf{F}_n^l\}$, we first apply linear projections to generate the query, key, and value matrices: $\mathbf{Q} =\phi_Q(\mathbf{F}_{in}),\mathbf{K} =\phi_K(\mathbf{F}_{in}),\mathbf{V} =\phi_V(\mathbf{F}_{in})$. The transformed features are then passed through a multi-head self-attention module followed by a two-layer MLP:
\begin{equation}
	\mathbf{F}_{out} = TwoLayerMLP(softmax(\frac{\mathbf{Q}\mathbf{K}^\textup{T}}{\sqrt{C}})\mathbf{V})
\end{equation}
By incorporating self-attention, each patch is able to perceive global contextual information, thereby improving the robustness of dictionary construction.
\par  
In the next subsection, we introduce the dictionary lookup process using individual patch features from $\mathbf{F}_q^l$. Specifically, we denote each patch feature as $\mathbf{x}_{q,p}^l \in \mathbb{R}^{1\times C}$, where $p=1,2,\cdots,HW$. The corresponding \textit{Dictionary Query} in $\mathbf{F}_Q^l$ is represented as $\mathbf{x}_{Q,p}^l$. For simplicity, we omit the subscript $p$ in the following discussion.

\subsection{Dictionary Lookup} 
Intuitively, if a query image patch contains an anomaly, its features $\mathbf{x}_{q}^l$ cannot be well retrieved from the \textit{Dictionary Value} $\mathbf{F}_V^l$. Based on this assumption, we propose three dictionary lookup strategies, each consisting of two main steps:
\par
1) \textit{Query-Key Matching}: The similarity $\mathbf{z}$ between the \textit{Dictionary Query} $\mathbf{x}_Q^l$ and the \textit{Dictionary Key} $\mathbf{F}_K^l$ is computed as: $ \mathbf{z} = \mathbf{x}_Q^l {\mathbf{F}^l}^\textup{T}_K$.
\par 
2) \textit{Weighted Fusion}: Using the computed similarity $\mathbf{z}$, the Retrieved Result $\mathbf{x}_r^l \in \mathbb{R}^{1\times C}$ is obtained by weighted fusion of the \textit{Dictionary Value} $\mathbf{F}_V^l$: $\mathbf{x}_r^l = \hat{\mathbf{w}} \mathbf{F}_V^l$, where $\hat{\mathbf{w}}\in \mathbb{R}^{1\times kHW}$ is a weight vector derived from $\mathbf{z}$, and is determined by one of the following strategies:
\begin{equation}
	\hat{\mathbf{w}} = \begin{cases}
		onehot(argmax(\mathbf{z})) & if\  \text{Maximun Lookup}, \\
		softmax(\mathbf{z}) & if \ \ \text{Dense Lookup}, \\
		SPM(\mathbf{z}) & if \ \text{Sparse Lookup}
	\end{cases}
\end{equation}
\par 
Specifically, for Maximum Lookup, the weight vector $\hat{\mathbf{w}}$ is a one-hot vector, selecting the \textit{Dictionary Value} with the highest similarity score. For Dense Lookup, the weight vector $\hat{\mathbf{w}}$ is obtained by applying the softmax function to $\mathbf{z}$, distributing the weights across all \textit{Dictionary Value} in a dense manner. For Sparse Lookup, we introduce a Sparse Probability Module (SPM), adapted from the approach proposed in \cite{sparse}, to sparsify $\mathbf{z}$ such that it emphasizes the most relevant  \textit{Dictionary Value} and reduces background redundancy as the number of reference images increases. We adopt Sparse Lookup as the default setting in this work.
\par  
\textbf{Sparse Probability Module.} The sparse fusion weight $\hat{\mathbf{w}}$ is computed by solving the optimization problem that sparsifies the similarity scores $\mathbf{z}$ and satisfies the probability simplex constraint:
\begin{equation}
	\operatorname*{arg\,min}_{\vartriangle} \frac{1}{2}\Vert \mathbf{w}- \mathbf{z}\Vert^2, \vartriangle = \{\mathbf{w}|\mathbf{w}_u \ge 0, \sum_{u=1}^{kHW}\mathbf{w}_u = 1\}
\end{equation}
where $\vartriangle$ is the probability simplex constraint. Solving the optimization problem yields $\hat{\mathbf{w}}_u = \max(\mathbf{z}_u-\tau, 0)$, where $\hat{\mathbf{w}}_u$ denotes the $u$-th element of $\hat{\mathbf{w}}$ and $\tau$ is a dynamic threshold determined by Algorithm~\ref{Alg1}. By repeating this process for all query patches, we obtain the final \textit{\textbf{Retrieved Result}} $\mathbf{F}_r^l$ under the sparse lookup strategy.
\begin{algorithm}[tb]
	\caption{The acquisition of the adaptive threshold $\tau$}
	\label{Alg1}
	\begin{algorithmic}[1] 
		\STATE  Sort $\mathbf{z}$ in descending order: $\mathbf{z}_1 \ge \mathbf{z}_2\ge \cdots \mathbf{z}_{kHW}$\\
		\STATE  Compute the cumulative sum: $Cum_{t} = \sum_{u=1}^{t}\mathbf{z}_t$ \\
		\STATE  Compute the candidate threshold: $\tau_t = (Cum_{t} - 1) / t$ \\
		\STATE Find the largest $t$ (denoted as $t^*$) satisfying $\mathbf{z}_t>\tau_t$, then the final threshold is $\tau = \tau_{t^*}$
	\end{algorithmic}
\end{algorithm}
\par 
\textbf{Query Loss.} How can the model acquire a class-generalizable dictionary lookup ability? Our answer is to use the \textit{\textbf{Query Feature}} $\mathbf{F}_q^l$ itself as a pseudo-label for self-supervised training. The core assumption is that normal patch features from $\mathbf{F}_q^l$ can always be retrieved from the constructed dictionary. To this end, we minimize the average distance $\mathbb{E}_{\mathcal{N}}[d]$ between the \textit{\textbf{Query Feature}} $\mathbf{F}_q^l$ and the \textit{\textbf{Retrieved Result}} $\mathbf{F}_r^l$ in normal regions. The query loss is defined as:
\begin{equation}
	\mathcal{L}_{q} = \sum_{l}\mathbb{E}_{\mathcal{N}}[d] = \sum_{l}\Bigl(\frac{1}{|\mathcal{N}|}\sum_{j\in \mathcal{N}}d(\mathbf{F}_{q,j}^l, \mathbf{F}_{r,j}^l )\Bigr)
\end{equation}
where $d(\mathbf{F}_{q,j}^l, \mathbf{F}_{r,j}^l )=1 - \langle\mathbf{F}_{q,j}^l,\mathbf{F}_{r,j}^l \rangle$ denotes the cosine distance between the $j^{\text{th}}$ patch, and $\langle\cdot  ,\cdot \rangle$ is the cosine similarity. The anomaly mask  $\mathbf{G} \in \{0,1\}^{HW}$ is obtained using the anomaly synthesis algorithm in DR{\AE}M \cite{Draem}. Here, $\mathcal{N} = \{j|\mathbf{G}_j = 0\}$ represents the index set of normal patches, and $|\mathcal{N}|$ is its cardinality.
 
\begin{table*}[t]
	\caption{Comparison with existing state-of-the-art methods under the 4-shot setting. The best results are highlighted in \textcolor{darkred}{red}, and the second-best results are marked in \textcolor{blue}{blue}. The symbol $\dag$ denotes methods based on CLIP, and (a,b,c) represents pixel-level (AUROC, PRO, AP). To ensure a fair comparison, all methods use the same normal reference images, and all CLIP-based methods employ the same backbone (ViT-L-14-336) and input resolution ($336\times 336$).} 
	\centering
	\label{Tab1}
	\renewcommand{\arraystretch}{1}
	\resizebox{1.8\columnwidth}{!}
	{
		\begin{tabular}{>{\centering\arraybackslash}p{2.2cm}*{5}{>{\centering\arraybackslash}p{2.7cm}}>{\centering\arraybackslash}p{2.9cm}>{\centering\arraybackslash}p{2.6cm}>{\columncolor{lightgreen}\centering\arraybackslash}p{2.6cm}}
			\toprule
			Datasets & \makecell[c]{RegAD \cite{RegAD} \\ (ECCV 22)}    & \makecell[c]{AnomalyGPT \cite{AnomalyGPT}\\(AAAI 24)} & \makecell[c]{FastRecon \cite{fastrecon}\\(ICCV 23)} & \makecell[c]{$\dag$ FastRecon+ \cite{fastrecon}\\ (ICCV 23)} & \makecell[c]{$\dag$ WinCLIP \cite{winclip}\\ (CVPR 23)}  & \makecell[c]{$\dag$ APRIL-GAN \cite{VAND}\\(CVPR 23)} & \makecell[c]{$\dag$ PromptAD \cite{PromptAD}\\(CVPR 24)} & \makecell[c]{$\dag$ DictAS\\(Ours)}             \\   \midrule
			\multicolumn{9}{c}{Industrial Datasets\quad (AUROC, PRO, AP)}                                                                                                                                                \\   \midrule
			MVTecAD \cite{mvtec} & (95.7, 86.0, 46.5) & (\textcolor{blue}{96.4}, 91.2, 52.9)   & (95.9, 79.9, 47.0)  & (96.3, 92.2, 53.9)   & (92.4, 83.8, 39.2) & (92.2, 86.6, 46.6)  & (96.0, \textcolor{blue}{92.4}, \textcolor{blue}{57.5}) & (\textcolor{darkred}{98.6}, \textcolor{darkred}{95.1}, \textcolor{darkred}{66.8}) \\
			VisA \cite{visa}     & (94.7, 72.8, 21.4) & (96.5, 65.4, 20.8)   & (96.0, 77.7, 31.1)  & (97.0, 86.2, 32.5)   & (96.0, 86.5, 25.7) & (96.2, 86.6, 30.6)  & (\textcolor{blue}{97.9}, \textcolor{blue}{89.5}, \textcolor{blue}{37.5}) & (\textcolor{darkred}{98.8}, \textcolor{darkred}{91.9}, \textcolor{darkred}{41.8}) \\
			MVTec3D \cite{mvtec3D}  & (96.9, 89.2, 13.3) & (96.6, 87.4, 27.8)   & (95.6, 83.6, 12.9)  & (97.1, 91.8, \textcolor{blue}{39.2})   & (96.6, 87.9, 24.0) & (96.4, 89.1, 33.1)  & (\textcolor{blue}{97.7}, \textcolor{blue}{92.1}, 36.9) & (\textcolor{darkred}{98.4}, \textcolor{darkred}{94.9}, \textcolor{darkred}{44.2}) \\
			MPDD \cite{MPDD}     & (94.9, 83.3, 16.4) & (\textcolor{blue}{97.7}, 93.2, \textcolor{blue}{40.8})   & (97.0, 87.5, 25.7)  & (97.4, 93.1, 37.8)   & (97.0, 90.7, 29.3) & (95.3, 86.9, 31.4)  & (97.3, \textcolor{blue}{94.0}, 40.5) & (\textcolor{darkred}{98.4}, \textcolor{darkred}{95.8}, \textcolor{darkred}{42.9}) \\
			BTAD \cite{BTAD}    & (97.3, 75.5, 44.1) & (96.2, 73.5, 50.6)   & (88.7, 62.1, 35.5)  & (\textcolor{blue}{97.4}, \textcolor{blue}{80.8}, 62.2)   & (90.3, 64.7, 28.5) & (93.3, 74.6, 50.9)  & (96.6, 80.1, \textcolor{blue}{62.5}) & (\textcolor{darkred}{98.0}, \textcolor{darkred}{83.3}, \textcolor{darkred}{66.8}) \\
			\textbf{Average}  & (95.9, 81.3, 28.3) & (96.7, 82.1, 38.6)   & (94.6, 78.2, 30.4)  & (97.0, 88.8, 45.1)   & (94.5, 82.7, 29.3) & (94.7, 84.8, 38.5)  & (\textcolor{blue}{97.1}, \textcolor{blue}{89.6}, \textcolor{blue}{47.0}) & (\textcolor{darkred}{98.4}, \textcolor{darkred}{92.2}, \textcolor{darkred}{52.5}) \\   \midrule
			\multicolumn{9}{c}{Medical Datasets \quad (AUROC, PRO, AP)}                                                                                                                                                   \\   \midrule
			RESC \cite{RESC}   & (87.9, 60.0, 18.1) & (86.7, 60.0, 28.5)   & (91.7, 71.7, 30.3)  & (95.8, 82.8, 68.5)   & (93.1, 75.7, 38.4) & (93.7, 77.6, 57.3)  & (\textcolor{blue}{96.8}, \textcolor{blue}{86.8}, \textcolor{blue}{71.3}) & (\textcolor{darkred}{97.5}, \textcolor{darkred}{89.7}, \textcolor{darkred}{74.9}) \\
			BrasTS \cite{BrasTS}   & (93.8, 70.2, 24.8) & (95.4, 73.6, 41.8)   & (92.5, 63.8, 24.0)  & (96.1, 73.8, 43.9)   & (93.3, 64.0, 33.4) & (91.3, 63.0, 40.0)  & (\textcolor{blue}{96.6}, \textcolor{blue}{77.0}, \textcolor{blue}{54.4}) & (\textcolor{darkred}{97.3}, \textcolor{darkred}{77.2}, \textcolor{darkred}{59.3}) \\
			\textbf{Average}  & (90.8, 65.1, 21.5) & (91.0, 66.8, 35.2)   & (92.1, 67.8, 27.1)  & (96.0, 78.3, 56.2)   & (93.2, 69.8, 35.9) & (92.5, 70.3, 48.7)  & (\textcolor{blue}{96.7}, \textcolor{blue}{82.2}, \textcolor{blue}{62.9}) & (\textcolor{darkred}{97.4}, \textcolor{darkred}{83.4}, \textcolor{darkred}{67.1})  \\ \bottomrule
		\end{tabular}
	}
\end{table*}

\subsection{Query Discrimination Regularization}
The query loss $\mathcal{L}_{q}$ enables the model learn how to retrieve from the dictionary and detect anomalies by measuring the distance between query features and their retrieved counterparts. However, this strong retrieval capability is a double-edged sword. If the model becomes excessively powerful in both feature extraction and retrieval, it may find combinations of normal features in the dictionary that closely matches any query, including anomalous ones. This weakens segmentation, as the distance can no longer clearly separate normal from abnormal regions. To mitigate this issue, we introduce two constraints: 
\par 
1) \textbf{Contrastive Query Constraint} enforces that the average distance between abnormal patches from $\mathbf{F}_q^l$ and their retrieved results from $\mathbf{F}_r^l$ is greater than that of normal patches, enhancing anomaly separability:
\begin{equation}
\mathcal{L}_{CQC} = \sum_{l}\max(0, \ \mathbb{E}_{\mathcal{N}}[d] - \mathbb{E}_{\mathcal{A}}[d])
\end{equation}
where $\mathbb{E}_{\mathcal{A}}[d]$ represents the average cosine distance between the \textbf{\textit{Query Feature}} and \textbf{\textit{Retrieved Result}} in abnormal regions, similar to $\mathbb{E}_{\mathcal{N}}[d]$ in Equation (7).
\par 
2) \textbf{Text Alignment Constraint} leverages CLIP’s text-image matching capability to enforce alignment between global retrieved results and the embedding space of normal text descriptions, preventing the retrieval of anomalous features from the dictionary. 
\par 
Inspired by WinCLIP \cite{winclip}, we adopt a two-class text prompt design to ensure effective CLIP-based alignment. Specifically, we construct a series of prompt templates and perform prompt ensembling by averaging the corresponding text embeddings. For example, the template \textquotedblleft a photo of a [state] [class]\textquotedblright, where [state] indicates an adjective describing normal or anomalous conditions (e.g., good / damaged), and [class] denotes the object category (e.g., wood, bottle).  The resulting text embedding is denoted as $\mathbf{F}_{text}\in \mathbb{R}^{2\times C}$. To obtain the global representation of the \textit{\textbf{Query Feature}}, $\mathbf{F}_q^l\in \mathbb{R}^{H\times W\times C}$ from different layers are concatenated along the channel dimension and global average pooling is applied over the spatial dimensions to obtain a compact representation. A linear layer is finally used to map the pooled features into $\mathbf{x}_q$ in the joint embedding space. The same process is applied to the \textit{\textbf{Retrieved Result}} $\mathbf{F}_r^l$ to yield the global feature $\mathbf{x}_r$. The final constraint is formulated as:
\begin{equation}
	\mathcal{L}_{TAC} = CE(\widetilde{\mathbf{x}}_r \widetilde{\mathbf{F}}_{text}^\textup{T}, 0) + CE(\widetilde{\mathbf{x}}_q\widetilde{\mathbf{F}}_{text}^\textup{T}, y_{q})
\end{equation}
where $CE(\cdot)$ represents the cross-entropy loss function \cite{Deep} and $y_{q}\in\{0,1\}$ is the image-level label derived from the synthesized query image. $\widetilde{(\ \cdot) \ }$ denotes the L2-normalized version along the embedding dimension. Details of the text prompt design are provided in Appendix A.3.

\subsection{Training and Inference}
\textbf{Training.} We train DictAS in a self-supervised manner, allowing any image $\mathbf{X}$ without pixel annotations as auxiliary data. The total loss function is defined as: 
\begin{equation}
	\mathcal{L} = \mathcal{L}_{q} + \lambda_1 \mathcal{L}_{CQC} + \lambda_2 \mathcal{L}_{TAC}
\end{equation}
where $\lambda_1$ and $\lambda_2$ are weighting coefficients for the regularization terms. Note that the number of simulated reference images is set to $k=1$ for training efficiency.
\par
\textbf{Inference.} During this stage, the dictionary is constructed using $k(k\ge 1)$ real normal reference images. The cosine distance between the query features and their retrieved counterparts is computed to generate the anomaly map. The final anomaly map $\mathbf{M}\in \mathbb{R}^{h\times w}$ is obtained by aggregating the maps from $L$ layers:
\begin{equation}
	\mathbf{M} = Up\Bigl(\frac{1}{2L}\sum_{l=1}^{L} \Bigl( 1 -\langle\mathbf{F}_{q}^l,\mathbf{F}_{r}^l \rangle\Bigl)\Bigl)
\end{equation}
where $Up(\cdot)$ represents the upsampling operation used to restore the original input resolution.

\begin{figure*}[t]
	\centering
	\includegraphics[width= 1.8\columnwidth]{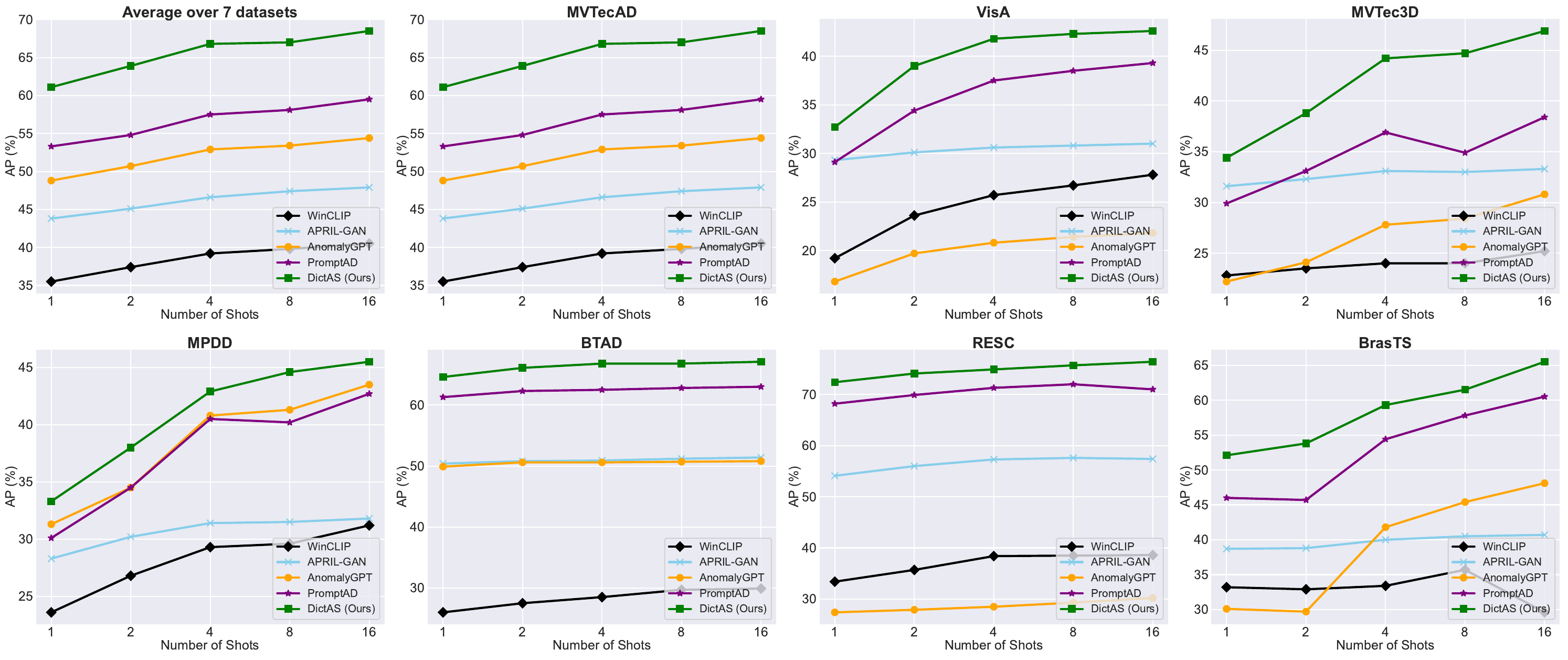}
	\caption{Comparison with four representative VLM-based methods under different numbers of shots. The pixel-level AP on seven datasets is reported. DictAS demonstrates consistent improvements over the compared methods for all shots. As the number of reference images increases, DictAS shows a greater advantage, unlike some methods that exhibit instability or even a decline (e.g., WinCLIP in BrasTS).}
	\label{all_shots}
\end{figure*}

\section{Experiments}
\subsection{Experimental Settings} 
\textbf{Datasets.} To evaluate the class-generalizable FSAS performance, experiments on seven real-world datasets from the industrial and medical domain are conducted. Specifically, five commonly used industrial anomaly segmentation datasets are adopted: MVTecAD \cite{mvtec}, VisA \cite{visa}, MVTec3D \cite{mvtec3D}, MPDD \cite{MPDD}, and BTAD \cite{BTAD}. Additionally, two medical datasets are included: BraTS \cite{BrasTS} for brain tumor segmentation and RESC \cite{RESC} for retinal lesion detection. For MVTec3D, only RGB images in the dataset are adopted. Following \cite{AnomalyGPT,ResAD,VAND}, we utilize all normal images from the industrial dataset VisA \cite{visa} as auxiliary data for self-supervised training and directly conduct few-normal-shot testing on remaining industrial and medical datasets. To evaluate VisA itself, MVTecAD \cite{mvtec} is adopted for auxiliary training. Notably, since a self-supervised paradigm is employed, any image without pixel-level annotations can serve as auxiliary training data, including those from natural scenes. In Appendix C.2, we further analyze the impact of different auxiliary training sets through ablation studies.
\par  
\textbf{Evaluation Metrics.} Since this paper primarily focuses on anomaly segmentation, we report three pixel-level metrics in the main text: Area Under the Receiver Operating Characteristic (AUROC), Per-Region Overlap (PRO), and the Average Precision (AP). As a supplement, we also report anomaly classification metrics in Appendix E.1, where the classification score for each image is obtained following the same strategy as APRIL-GAN \cite{VAND}. We present the performance for different values of the few-shot normal samples, with $k$ set to 1, 2, 4, 8, and 16.
\par 
\textbf{Implementation Details.} Similar to recent state-of-the-art FSAS methods \cite{winclip, VAND, PromptAD}, we adopt the CLIP model (ViT-L-14-336), pretrained by OpenAI \cite{CLIP}, as the default backbone in DictAS. All input images are uniformly resized to $336 \times 336$ before being fed into the model. During training, we extract the 6th, 12th, 18th, and 24th layers from the frozen image encoder as patch-level features, following APRIL-GAN \cite{VAND}. This multi-level feature selection balances low-level appearance and high-level semantics, and facilitates a fair comparison with prior work. The balancing coefficients $\lambda_1$ and $\lambda_2$ for the regularization loss are both set to 0.1 by default. During the auxiliary training stage, two types of data transformations—Geometric Transformations (e.g., Random Rotation) and Occlusion Augmentations (e.g., Random GridDropout)—are applied to the raw images to generate reference images. We use the Adam optimizer \cite{Adam} to train DictAS for 30 epochs, with an initial learning rate of 0.0001 and a batch size of 24. All experiments were conducted on a single NVIDIA 3090 with 24GB of GPU memory. More details can be found in Appendix A.
 \begin{figure}[t]
	\centering
	\includegraphics[width= 1\columnwidth]{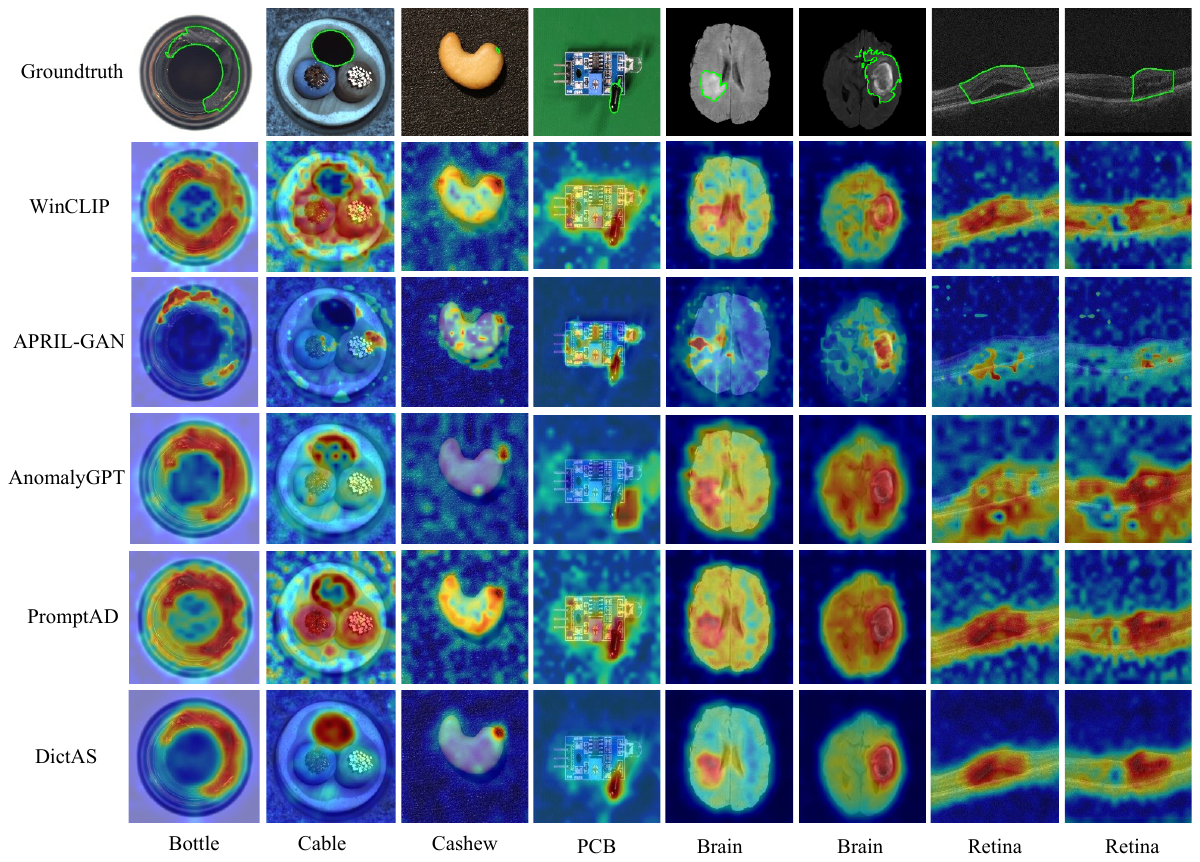}
	\caption{Qualitative comparison of anomaly segmentation results across different FSAS methods.}
	\label{Fig4}
\end{figure}
\begin{table}[t]
	\caption{The comparison of model efficiency on MVTecAD under the 4-shot setting (mean\dev{std} for AP).} 
	\centering
	\label{Efficiency}
	\renewcommand{\arraystretch}{1}
	\resizebox{0.95\columnwidth}{!}
	{
		\begin{tabular}{cccccc}
			\toprule
			Method     & Backbone       & Resolution & \makecell[c]{GPU Cost\\(GB)} & \makecell[c]{Time\\(ms)} & AP (\%)       \\  \midrule
			RegAD \cite{ResAD}      & ResNet18       & $224\times224$        & 7.1           & 5790.1    & 46.5\dev{1.2} \\
			AnomalyGPT \cite{AnomalyGPT}& ImageBind-Huge & $224\times224$        & 19.1          & 1555.2    & 52.9\dev{0.5} \\
			FastRecon \cite{fastrecon} & Wide-ResNet50  & $336\times336$        & \textbf{1.0}           & 110.9     & 47.0\dev{0.7} \\
			FastRecon+ \cite{fastrecon} & ViT-L-14-336   & $336\times336$        & 3.4           & 150.5     & 53.9\dev{0.6} \\
			WinCLIP \cite{winclip}   & ViT-L-14-336   & $336\times336$        & 15.8          & 8227.5    & 39.2\dev{1.1} \\
			APRIL-GAN \cite{VAND} & ViT-L-14-336   & $336\times336$        & 3.8           & 227.6     & 46.6\dev{0.5} \\
			PromptAD \cite{PromptAD}  & ViT-L-14-336   & $336\times336$        & 2.1           & 81.3      & 57.5\dev{0.6} \\  \midrule \rowcolor{lightgreen}
			DictAS     & ViT-L-14-336   & $336\times336$        & 3.4           & \textbf{73.5}      & \textbf{66.8\dev{0.4}} \\  \bottomrule
		\end{tabular}
	}
\end{table}
\subsection{Comparison with State-of-the-art methods}
\textbf{Competing Methods.} In this study, we compare our DictAS with six state-of-the-art (SOTA) methods: RegAD \cite{RegAD}, AnomalyGPT \cite{AnomalyGPT}, FastRecon \cite{fastrecon}, WinCLIP \cite{winclip}, APRIL-GAN \cite{VAND}, and PromptAD \cite{PromptAD}. Among these, PromptAD follows a class-dependent FSAS setting, while the other methods adopt a class-generalizable setting. To ensure a fair comparison, all methods were evaluated using the same $k$ normal reference images and repeated five times. Additionally, we introduce a variant of FastRecon that incorporates CLIP image encoder as the feature extractor, referred to as FastRecon+. All CLIP-based methods \cite{winclip, VAND, PromptAD} utilize the same backbone and input image resolution as our DictAS. 
\par 
\textbf{Quantitative Comparison.} Table \ref{Tab1} presents the quantitative results of ZSAS across different datasets under the 4-shot setting in both industrial and medical domains. DictAS consistently outperforms existing methods, achieving state-of-the-art FSAS performance. Compared to the suboptimal method, DictAS improves AUROC, PRO, and AP by 1.3\%, 2.6\% and 5.5\%, respectively, on industrial datasets, and by 0.7\%, 1.2\%, 4.2\% on medical datasets. Notably, most of the suboptimal results are achieved on PromptAD, which is a class-dependent method that requires fine-tuning a new model for each unseen class, limiting its scalability. In contrast, our DictAS removes this dependency, enabling class-generalizable FSAS in a unified model, which is more flexible and efficient. In addition, DictAS outperforms other class-generalizable methods (e.g., WinCLIP, APRIL-GAN) by a large margin across all metrics. This is because it adopts a self-supervised training paradigm to learn cross-category dictionary lookup capabilities, rather than relying on prior knowledge from previous real anomaly samples or pretraining. Figure \ref{all_shots} further illustrates the variations in AP across seven datasets under different numbers of shots. It can be observed that DictAS demonstrates consistent improvements over the compared methods for all shots. This improvement is attributed to our sparse lookup strategy, which helps reduce feature redundancy at higher shots.
\par 
\textbf{Qualitative Comparison.} In Figure \ref{Fig4}, we present visualization results from industrial and medical datasets under the 4-shot setting. Overall, our DictAS achieves more precise and complete anomaly localization, attributed to the global lookup capability learned during self-supervised training via the designed \textit{Dictionary Key/Query/Value Generator}.

\par 
\textbf{Efficiency Comparison.} Table \ref{Efficiency} compares the efficiency of different methods, including the average inference time per image and the maximum GPU memory usage per image during inference. The experimental results show that the proposed DictAS achieves the fastest inference speed and the best FSAS performance. Although few-shot methods are susceptible to variations in reference images, experimental results show that DictAS exhibits greater stability with lower standard deviation compared to other methods. 

\subsection{Ablation}
Unless otherwise specified, all ablation experiments in this subsection are conducted on the MVTecAD \cite{mvtec} dataset under the 4-shot setting. 
\par
\textbf{The effects of components.} Ablation studies on both modules and loss functions are conducted as shown in Table \ref{Tab3}. In the module ablation, it can be observed that when any generator is removed and the original image features are employed instead, the performance of DictAS drops to varying degrees. This is mainly due to the fact that the designed \textit{AttnBlock} in the generator provides global attention for dictionary lookup, which facilitates sparse matching of \textit{Query}-\textit{Key} pairs and weighted fusion of the \textit{Dictionary Value}. For the loss function, we performed ablation on the two regularization constraints and found that they jointly influence the final FSAS performance. Compared to the constraint $\mathcal{L}_{CQC}$, which affects the global feature space, constraint $\mathcal{L}_{TAC}$ enhances the discriminability of anomalous regions in the fine-grained feature space, resulting in greater performance gains (e.g. $2.2\%\uparrow$AP vs. $1.8\%\uparrow$AP).
\par
\begin{figure}[t]
	\centering
	\includegraphics[width= 0.9\columnwidth]{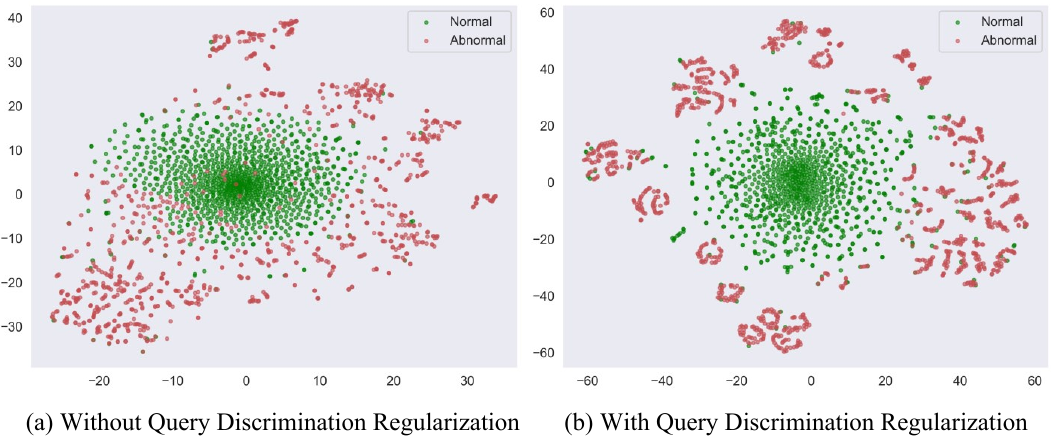}
	\caption{Feature t-SNE visualization. (a) / (b) represent the cases without / with Query Discrimination Regularization, respectively.}
	\label{Fig5}
\end{figure}

 
 To further visualize the impact of the proposed Query Discrimination Regularization on the results, we randomly selected 40 images from the category \textit{cable} of MVTecAD and performed t-SNE visualization on the feature map $(\mathbf{F}_r - \mathbf{F}_q)$, which is obtained by taking the residual between the \textbf{\textit{Query Feature}} and \textbf{\textit{Retrieved Result}} from the same layer. As shown in Figure \ref{Fig5}, without the Query Discrimination Regularization ( w/o $\mathcal{L}_{CQC}$ and $\mathcal{L}_{TAC}$), the boundary between normal and anomalous regions in the residual features remains ambiguous. However, when the regularization is applied, the residual features become more distinguishable between normal and anomalous regions, thereby facilitating easier anomaly discrimination.
 
\begin{table}[t]
	\caption{Ablation on different components (\%).} 
	\centering
	\label{Tab3}
	\renewcommand{\arraystretch}{1}
	\resizebox{0.9\columnwidth}{!}
	{
	\begin{tabular}{>{\centering\arraybackslash}p{1.2cm}>{\centering\arraybackslash}p{3.1cm}*{3}{>{\centering\arraybackslash}p{1cm}}}
		\toprule
		&                     & AUROC &  PRO & AP   \\   \midrule
		\multirow{3}{*}{\makecell[c]{Module \\ Ablation}}  & w/o \textit{Query Generator} & 97.5  & 94.2  & 63.5 \\
		& w/o \textit{Key Generator}   & 97.9  & 94.5  & 63.8 \\
		& w/o \textit{Value Generator} & 98.0  & 94.6  & 64.2 \\   \midrule
		\multirow{3}{*}{\makecell[c]{Loss \\ Ablation}}    & w/o  $\mathcal{L}_{CQC}$             & 97.4  & 94.1  & 64.6 \\
		& w/o $\mathcal{L}_{TAC}$              & 98.0  & 94.6  & 65.0 \\
		& w/o $\mathcal{L}_{CQC}$ and $\mathcal{L}_{TAC}$       & 97.1  & 93.5  & 63.7 \\  \midrule
		\rowcolor{lightgreen}
		& DictAS              & \textbf{98.6}  & \textbf{95.1}  & \textbf{66.8}  \\  \bottomrule
		\end{tabular}
}
\end{table}

\begin{table}[t]
	\caption{Ablation on dictionary lookup strategies. The pixel level AP (\%) is employed to evaluate under different shots.} 
	\centering
	\label{Tab4}
	\renewcommand{\arraystretch}{1}
	\resizebox{0.9\columnwidth}{!}
	{
		\begin{tabular}{>{\centering\arraybackslash}p{2.8cm}*{5}{>{\centering\arraybackslash}p{1.1cm}}}
			\toprule
			Lookup   Strategy & 1-shot & 2-shot & 4-shot & 8-shot & 16-shot \\   \midrule 
			Maximum Lookup    & 52.2   & 56.5   & 59.1   & 59.7   & 60.6    \\
			Dense Lookup      & 60.2   & 62.9   & 63.7   & 63.6   & 63.8    \\  \rowcolor{lightgreen}
			Sparse Lookup     & \textbf{61.1}   & \textbf{63.9}   & \textbf{66.8}   & \textbf{67.0}   & \textbf{68.5}   \\  \bottomrule
		\end{tabular}   
	}
\end{table}
\textbf{The effects of dictionary lookup strategies.} Table \ref{Tab4} presents an evaluation of various dictionary lookup strategies. The experimental results indicate that the proposed Sparse Lookup achieves the highest AP across all shot settings. Moreover, compared to other strategies, Sparse Lookup demonstrates greater advantages when the number of reference images increases (e.g., in the 8-shot and 16-shot settings). This is primarily because it maintains query sparsity while utilizing a larger set of Dictionary Values, effectively mitigating feature redundancy in the dictionary. 

\textbf{The effects of backbone and resolution.} Table \ref{Tab5} evaluates the impact of different input resolutions and CLIP backbones on DictAS. Despite the significant differences among the pre-trained backbones (top three rows), the anomaly segmentation performance of DictAS remains relatively stable, especially in terms of AUROC. This indicates that our method is not highly dependent on the original visual representations. The reason is that self-supervised training on the auxiliary dataset equips the model with category-agnostic dictionary lookup capabilities, thereby reducing reliance on original features. Additionally, with the increase in input resolution, we observe a significant improvement in AP, but the inference time per image also increases considerably. To strike a balance, we adopt a default setting of $336^2$ resolution with the ViT-L-14-336 backbone.

\begin{table}[t]
	\caption{Ablation on backbone and input resolution (\%).} 
	\centering
	\label{Tab5}
	\renewcommand{\arraystretch}{1}
	\resizebox{0.9\columnwidth}{!}
	{
	\begin{tabular}{>{\centering\arraybackslash}p{2.2cm}>{\centering\arraybackslash}p{1.5cm}*{3}{>{\centering\arraybackslash}p{1cm}}>{\centering\arraybackslash}p{1.5cm}}
	\toprule
		Backbone     & Resolution & AUROC & PRO & AP   & Time (ms) \\  \midrule
		ViT-B-16-224 & $336\times336$        & 98.1  & 93.3  & 64.8 & 43.3     \\
		ViT-L-14-224 & $336\times336$        & 98.3  & 94.3  & 66.2 & 73.5     \\  \rowcolor{lightgreen}
		ViT-L-14-336 & $336\times336$        & 98.6  & 95.1  & 66.8 & 73.5     \\
		ViT-L-14-336 & $420\times420$        & 98.6  & 95.4  & 67.6 & 130.2     \\
		ViT-L-14-336 & $518\times518$        & \textbf{98.7}  & \textbf{95.6}  & \textbf{68.7} & 235.6     \\  \bottomrule
		\end{tabular}   
}
\end{table}

\section{Conclusion}
In this paper, we introduce DictAS, a novel framework for class-generalizable few-shot anomaly segmentation (FSAS). Inspired by human inspectors, we reformulate FSAS as a dictionary lookup task. An anomaly is detected when the query feature cannot be retrieved from the dictionary constructed from normal reference image features. Through self-supervised training on seen classes from an auxiliary dataset, DictAS learns a transferable dictionary lookup ability, enabling it to generalize effectively to unseen classes in FSAS. To further enhance anomaly discrimination, we introduce query discrimination regularization, which is jointly optimized with the query loss to make anomalous features less retrievable from the dictionary. The final anomaly map is computed based on the cosine distance between the query feature and its retrieval result. Extensive experiments on seven industrial and medical datasets demonstrate that DictAS surpasses SOTA methods in FSAS performance while also achieving the fastest inference speed with comparable stability.

\section*{\textbf{Acknowledgement}}
This work is supported in part by the National Natural Science Foundation of China under Grant Nos. 62373350 and 62371179; in part by the Youth Innovation Promotion Association CAS (2023145); in part by the Beijing Nova Program 20240484687; in part by the Beijing Municipal Natural Science Foundation (China) 4252053; in part by the Longmen Laboratory \textquotedblleft \textit{Research and Development Project of General-Purpose AI Platform Software Based on Industrial Quality Inspection Large Model}\textquotedblright.

{
	\small
	\bibliographystyle{ieeenat_fullname}
	\bibliography{main}
}
\clearpage
\onecolumn
\appendix
\begin{center}
	\large \textbf{Appendix for DictAS: A Framework for Class-Generalizable Few-Shot Anomaly Segmentation via Dictionary Lookup}
\end{center}
\par
This appendix includes the following five parts: 1) More experimental details (e.g. datasets, self-supervised training) in Section \ref{sec1}; 2) Detailed description of SOTA methods and comparison with contemporaneous approaches (e.g., MetaUAS, ResAD) in Section \ref{sec2}; 3) Additional ablation studies (e.g., hyperparameters, auxiliary datasets, data transformations) in Section \ref{sec3}; 4) Limitations of our methods in Section \ref{sec4}; 5) Presentation of more detailed quantitative and qualitative results of few-shot anomaly classification / segmentation in Section \ref{sec5}.
\section{Experimental Details}
\label{sec1}
\subsection{Details of the Datasets}
\begin{table*}[h]
\caption{Key statistics of the datasets. $(a,b)$ in the training/testing sets denotes the number of normal and abnormal samples, respectively. $|\mathcal{C}|$ is the number of categories. Note that anomaly segmentation datasets have only normal images in the training set.}
	\centering
	\label{Dataset}
	\renewcommand{\arraystretch}{1.2}
	\resizebox{0.8\columnwidth}{!}
	{
		\begin{tabular}{>{\centering\arraybackslash}p{2.2cm}>{\centering\arraybackslash}p{2.4cm}>{\centering\arraybackslash}p{2.2cm}>{\centering\arraybackslash}p{2.5cm}>{\centering\arraybackslash}p{2.2cm}>{\centering\arraybackslash}p{2.2cm}>{\centering\arraybackslash}p{2.2cm}>{\centering\arraybackslash}p{4cm}}
			\toprule
			Domain                      & Dataset & Category      & Modality          & $|\mathcal{C}|$  & Testing Set & Training Set & Usage                       \\  \midrule
			\multirow{5}{*}{Industrial} & MVTecAD \cite{mvtec} & Obj \&texture & Photography       & 15 & (467, 1258) & (3629, 0)  & Industrial defect detection \\
			& VisA \cite{visa}   & Obj           & Photography       & 12 & (962, 1200) & (8659, 0)  & Industrial defect detection \\
			& MVTec3D \cite{mvtec3D} & Obj           & Photography+Depth & 10 & (249, 948)  & (2656, 0)  & Industrial defect detection \\
			& MPDD \cite{MPDD}   & Obj           & Photography       & 6  & (176, 282)  & (888, 0)   & Industrial defect detection \\
			& BTAD \cite{BTAD}   & Obj           & Photography       & 3  & (451, 290)  & (1799, 0)  & Industrial defect detection \\  \midrule
			\multirow{2}{*}{Medical}    & RESC \cite{RESC}   & Retina           & Photography       & 1  & (1041, 764) & (4297, 0)  & Retinal Lesion Detection    \\
			& BrasTS \cite{BrasTS} & Brain           & Radiology(MRI)    & 1  & (828, 1948) & (4211, 0)  & Brain Tumor Segmentation   \\  \midrule
		\end{tabular}
	}
\end{table*}
In this study, we conduct extensive experiments on 7 public datasets covering  industrial and medical domains to assess the effectiveness of our methods, including MVTecAD \cite{mvtec}, VisA \cite{visa}, MVTec3D \cite{mvtec3D}, MPDD \cite{MPDD}, BTAD \cite{BTAD}, RESC \cite{RESC} and  BrasTS \cite{BrasTS}. The key statistics for these datasets are demonstrated in Table \ref{Dataset}. In this study, normal reference images are randomly selected from the training set, and all samples from the testing set are used to evaluate the model's performance. By default, all samples in the VisA training set are treated as seen classes for self-supervised training and are subsequently tested on other datasets. For VisA itself, the training set in MVTeAD is used as an auxiliary training dataset.
\subsection{Details of Self-Supervised Training}
This subsection further elaborates on the online construction of auxiliary data for self-supervised training.
\par
\begin{figure}[t]
	\centering
	\includegraphics[width= 0.7\columnwidth]{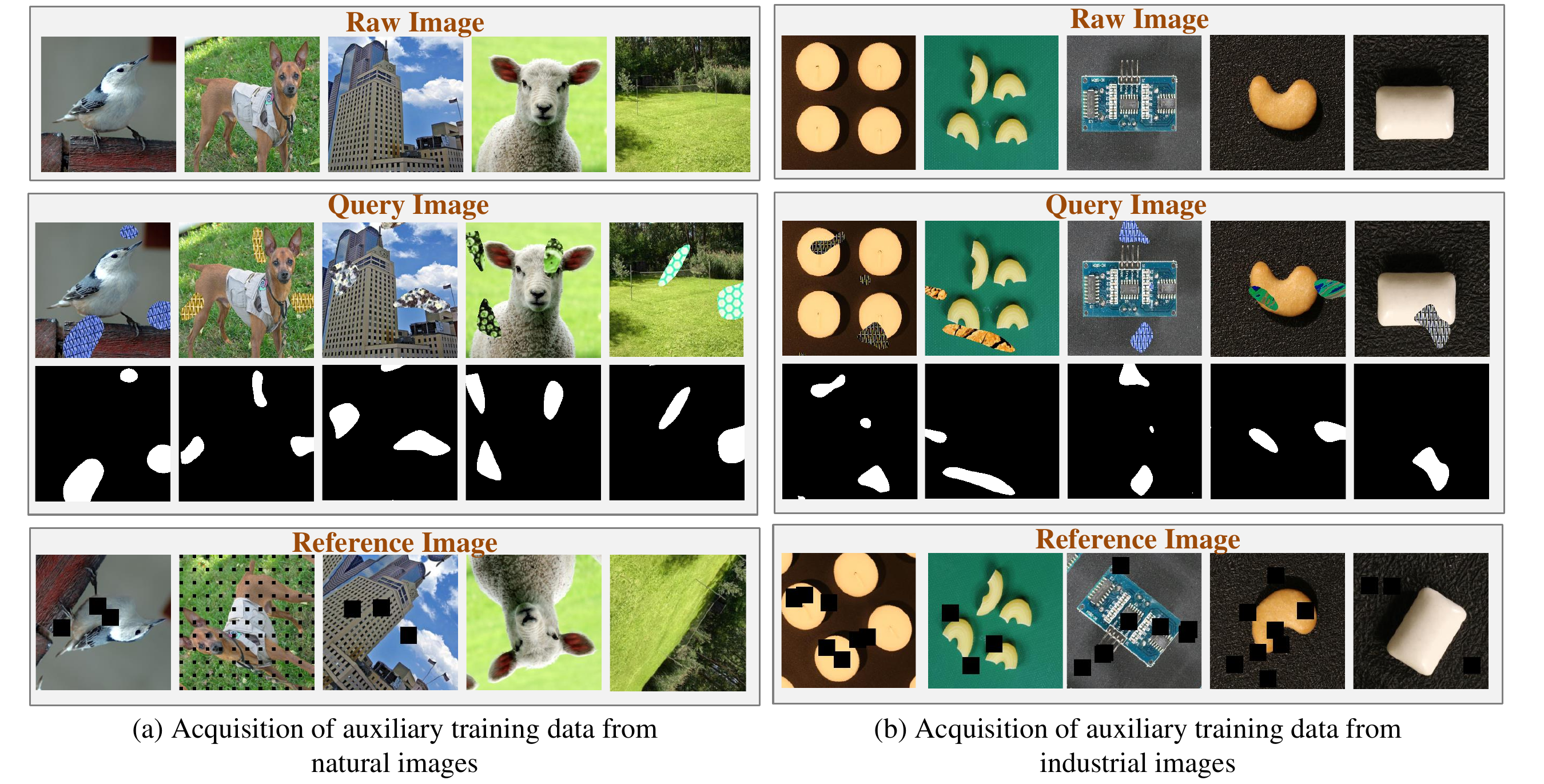}
	\caption{\textbf{Acquisition of the auxiliary training data.} Given a raw image without pixel-level annotations, the query image is generated using an anomaly synthesis algorithm \cite{Draem}, while the normal reference image is obtained via data transformations (e.g., random rotation). Both natural images shown in (a) and industrial images shown in (b) can be utilized as sources to construct auxiliary training data.} 
	\label{aug_two}
\end{figure}

In the self-supervised training stage,both query and reference images are dynamically constructed from raw images belonging to any seen class. Note that this process is conducted online. Specifically, given a raw image $\mathbf{X}$, we apply random transformations (e.g., random rotation) to generate a corresponding reference image, simulating the few normal reference images $\mathbf{X}_n$ available in the real anomaly segmentation process. In DictAS, we by default use Geometric Transformations and Occlusion Transformations as shown in Figure \ref{aug_two}. Detailed descriptions and parameters for each transformation type are provided in Listing 1. Additional ablation studies investigating the effect of different transformation types can be found in Section C.
\par 
For the query image $\mathbf{X}_q$, it is derived from the raw image using the anomaly synthesis algorithm proposed in DRAEM \cite{Draem}. The detailed strategy for synthesizing the query image during self-supervised training is described in Algorithm A. Alongside the synthesized image, the pixel-level pseudo-label $\mathbf{G}$ and the image-level pseudo-label $y_q$ are also generated using the Berlin noise mask. These pseudo-labels are used to compute the query contrastive loss and the text alignment loss, both of which act as regularization terms during self-supervised training.
\newpage
\begin{flushleft}
	\begin{lstlisting}[caption=Data transformation for generating the reference image in the self-supervised training stage., label=lst:augmentations, language=Python]
import albumentations as A
import cv2

img_trans_for_reference = A.Compose([
	A.RandomRotate90(p = 1),
	A.Rotate(limit=[30, 270], p=1.0),
	A.HorizontalFlip(p=0.5),
	A.VerticalFlip(p=0.5),
	A.GridDropout(ratio=0.3, p=0.5),
	A.CoarseDropout(max_holes=8, max_height=32, max_width=32, p=0.5),
	], is_check_shapes=False)
X_raw = cv2.imread("raw_img_path")   # Read the raw image
# Perform data transformation on the raw image to simulate the reference image.
X_reference = img_trans_for_reference(img = X_raw)  
	\end{lstlisting}
\end{flushleft}

\begin{algorithm}[h]
	\caption{Anomaly synthesis strategy for generating the query image in the self-supervised training stage.}
	\label{Alg2}
	\textbf{Input}: Raw image $\mathbf{X}$;  Anomaly source image $\mathbf{A}$; Perlin
	noise generator $P$;  Image size $H$ and $W$; Noise resolution $r_x$ and $r_y$; Blending parameter $\gamma$; Binarization threshold $\lambda$ \\
	\textbf{Output}: Query image $\mathbf{X}_q$, pixel-level pseudo-label $\mathbf{G}$, image-level pseudo-label $y_q$.
	\begin{algorithmic}[1] 
	\WHILE{True}
	\STATE $\mathbf{G}$ $\leftarrow$ where($P(H,w,r_x,r_y)>\lambda$)
	 \STATE $\mathbf{M}_A$ $\leftarrow$ $G$ $\times$ $\mathbf{X}$
	\STATE $\overline{\mathbf{M}}_A$ $\leftarrow$  1 - $\mathbf{M}_A$
	\STATE  $\mathbf{X}_q$ $\leftarrow$ $\gamma(\mathbf{M}_A \odot \mathbf{A}) + (1-\gamma)(\mathbf{M}_A \odot \mathbf{X})+ \overline{\mathbf{M}}_A \odot \mathbf{X}$ 
	\ENDWHILE
	\IF {SUM($\mathbf{G}$) is 0}
	\STATE $y_q$ $\leftarrow$ 0
	\ELSE 
	\STATE $y_q$ $\leftarrow$ 1
	\ENDIF
	\STATE return $\mathbf{X}_q$, $\mathbf{G}$, $y_q$
	\end{algorithmic}
\end{algorithm}
\subsection{Details of Text Prompt Design}

\begin{figure*}[h]
	\centering
	\includegraphics[width=0.9\columnwidth]{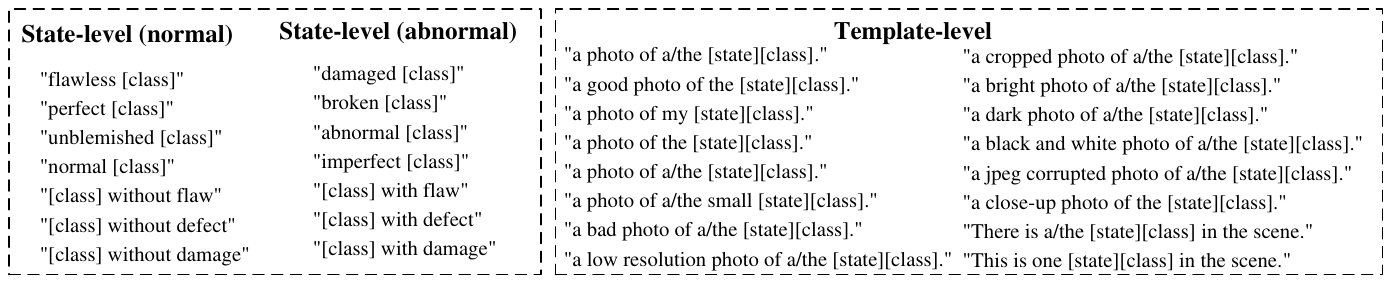}
	\caption{Detailed design of prompt template and normal/abnormal [state] words for text prompt ensemble.}
	\label{text_prompt}
\end{figure*}

In this work, two types of text prompts (normal descriptions and anomaly descriptions) are fed into the text encoder of CLIP to generate text embeddings. The global image representation obtained from the Retrieved Result $\mathbf{F}_r^l$ is constrained to align with the normal text embedding space, thereby enhancing anomaly discrimination capability. Since the design of text prompts is not the focus of this study, we directly follow the design principles of WinCLIP \cite{winclip} (i.e. text prompt ensemble). Specifically, to obtain normal text embeddings, the object category name (e.g., bottle) and state are inserted into predefined prompt templates to generate multiple semantically similar normal prompts. These prompts are encoded by the text encoder, and the resulting embeddings are averaged to form the final normal text representation. Similarly, the abnormal text embeddings are constructed in the same manner by replacing the state with an anomalous one. The details of the prompt template and the settings of normal/abnormal [state] are illustrated in Figure~\ref{text_prompt}.

\subsection{Details of Implementation}
Similar to recent state-of-the-art FSAS methods \cite{winclip, VAND, PromptAD}, we adopt the CLIP model (ViT-L-14-336), pretrained by OpenAI \cite{CLIP}, as the default backbone for our DictAS. All input images are uniformly resized to $336 \times 336$ before being fed into the model. During training, we extract the 6th, 12th, 18th, and 24th layers from the frozen image encoder as patch-level features similar to \cite{VAND}. To increase the receptive field, average pooling with a kernel size of 3 is applied to the patch-level features extracted from the CLIP image encoder. The regularization loss balancing coefficients, $\lambda_1$ and $\lambda_2$, are both set to 0.1 by default. During the auxiliary training phase, two types of data transformations—Geometric Transformations (e.g., Random Rotation) and Occlusion Augmentations (e.g., Random GridDropout)—are applied to the raw images to generate reference images. For computational efficiency, the number of reference images is set to $k=1$ during training. During inference, $k\geq 1$ normal reference images are used as visual prompts. To ensure a fair comparison, all methods are evaluated using the same $k$ normal reference images. Each experiment is repeated five times with different random seeds. DictAS is trained for 30 epochs using the Adam optimizer \cite{Adam}, with an initial learning rate of 0.0001 and a batch size of 24. All experiments are conducted on a single NVIDIA RTX 3090 GPU with 24 GB of memory.
\par

\section{State-of-the-art Methods}
\label{sec2}
\subsection{Method Introduction and Comparison Details}
\begin{itemize}
	
	\item \textbf{WinCLIP} \cite{winclip} is one of the earliest works based on CLIP for the zero/few shot anomaly segmentation task. Since the vanilla CLIP \cite{CLIP} does not align text with fine-grained image features during pretraining, it addresses this limitation by dividing the input image into multiple sub-images using windows of varying scales. The final language-guided anomaly segmentation results are derived by harmoniously aggregating the classification outcomes of sub-images corresponding to the same spatial locations. To leverage the few normal reference images, it also employs memory bank-based nearest neighbor retrieval to obtain visually guided anomaly maps. For a fair comparison, we report the results using ViT-L-14-336 as the backbone with an input resolution of $336\times 336$, based on the reproduced code from \cite{AnomalyCLIP}.

	\item \textbf{APRIL-GAN} \cite{VAND} adopts the handcrafted textual prompt design strategy from WinCLIP. However, for aligning textual and visual features, it introduces a linear adapter layer to project fine-grained patch features into a joint embedding space. After being trained on real anomalous samples with pixel-level, it can directly generalize to unseen classes. A memory-bank strategy like WinCLIP \cite{winclip} is also adopted to enhance text-image alignment results. For a fair comparison, we retrained the model using the official code on ViT-L-14-336 with a resolution of $336\times 336$ and re-evaluated it across all industrial and medical datasets. 
	
	\item \textbf{RegAD} \cite{RegAD} first proposed a feature registration strategy using a spatial transformer network for class-generalizable FSAS. With a meta-learning training approach, it demonstrates strong generalization to unseen classes. However, its performance on unseen category objects heavily depends on extensive augmentation of normal reference images and utilizes distribution estimation to generate the final anomaly map, making it less memory-efficient. In this work, we retrained RegAD using the same auxiliary dataset as our DictAS, i.e., trained on all classes of the VisA training set and tested on other datasets. For evaluation on VisA, the weights were obtained using MVTecAD as the auxiliary training set. Since RegAD has a specific backbone-dependent network structure, the backbone and resolution from the original paper were adopted (ResNet-18, $224\times 224$).
	
	\item \textbf{Fastrecon} \cite{fastrecon} models class-generalizable FSAS as a feature reconstruction problem based on linear regression. By designing a distribution regularization term and solving the analytical solution, it demonstrates excellent cross-category generalization in a training-free manner. However, as the number of reference images increases, the linear model may theoretically overfit arbitrary features, which means that Fastrecon still faces the challenge of over-reconstruction. In this version, we used the official code and backbone (wide-resnet50, $336\times$336) from the original paper and tested it across all datasets.
	
	\item \textbf{Fastrecon+} \cite{fastrecon} is a reimplementation of Fastrecon, utilizing the CLIP image encoder as the feature extractor. For a fair comparison, ViT-L-14-336 with a resolution of $336 \times 336$ is adopted. Following their original paper, we extracted the two intermediate patch-level features (the 12th and 18th layers) and concatenated them along the embedding dimension to construct new features. The other experimental hyperparameters are set to be the same as those in the original paper.
	
	\item \textbf{AnomalyGPT} \cite{AnomalyGPT} is a class-generalizable FSAS method that integrates a large language model for anomaly segmentation and supports multi-turn dialogues with users. It employs supervised training using synthetic anomaly data, allowing the model to generalize to new classes. We conducted experiments using the official code and evaluated the model’s FSAS performance in the same way as our DictAS. To use the officially pre-trained weights, the original backbone and input image resolution were adopted (ImageBind-Huge, $224\times 224$).
	
	\item \textbf{PromptAD} \cite{PromptAD} is a class-dependent FSAS method, which is different from other CLIP-based approaches. It directly trains on normal reference images for each class and evaluates on the test set of the same object category. Moreover, it proposes a one-class prompt learning
	method for few-shot anomaly segmentation. Although it outperforms most FSAS methods, the need for fine-tuning on each category limits its practicality in scenarios involving data privacy or rapidly changing environments. For fairness in comparison, we retrained the model on ViT-L-14-336 using an input resolution of $336 \times 336$.
	
    \item \textbf{MetaUAS} \cite{MetaUAS} proposes viewing FSAS as a segmentation change problem. By leveraging meta-learning training on a synthetic dataset, it enables the acquisition of a universal model capable of detecting anomalies in unseen classes. However, it is only applicable to situations where a single normal sample is used as the visual prompt (i.e., 1-shot). In this paper, we use it as a concurrent method and compare it with our DictAS.
    
    \item \textbf{ResAD} \cite{ResAD} proposes using learned residual feature distributions to reduce feature variations across different classes for class-generalizable FSAS. It ultimately transforms the anomaly segmentation problem into an out-of-distribution detection problem using a Feature Distribution Estimator, achieving strong performance on unseen classes. In this paper, we employ it as a concurrent method and compare its performance with our DictAS.

\end{itemize}
\subsection{Comparison with Concurrent Methods}

\begin{table*}[h]
	\caption{Comparison with the concurrent state-of-the-art methods. The pixel-level AUROC (\%) is reported, and the best results are highlighted in \textbf{bold}. The experimental results of MetaUAS and ResAD are taken from their original papers. }
	\centering
	\label{SOTA}
	\renewcommand{\arraystretch}{1.2}
	\resizebox{0.8\columnwidth}{!}
	{
	\begin{tabular}{>{\centering\arraybackslash}p{2.0cm}>{\centering\arraybackslash}p{2.2cm}>{\centering\arraybackslash}p{3.0cm}>{\centering\arraybackslash}p{2.2cm}*{4}{>{\centering\arraybackslash}p{2.0cm}}}
		\toprule
		&         & Backbone        & MVTecAD \cite{mvtec} & VisA \cite{visa} & BTAD \cite{BTAD} & MVTec3D \cite{mvtec3D} & BrasTS \cite{BrasTS} \\  \midrule
		\multirow{3}{*}{1-shot} & MetaUAS \cite{MetaUAS} & EfficientNet-b4 & 94.6    & 92.2 & ---    & ---       & ---      \\
		& DictAS  & ViT-B-16        & 97.1    & 97.3 & ---    & ---       & ---      \\
		& DictAS  & ViT-L-14-336    & \textbf{97.7}    & \textbf{98.0} & \textbf{97.4} & \textbf{97.5}    & \textbf{96.5}   \\  \midrule
		\multirow{2}{*}{2-shot} & ResAD \cite{ResAD}  & ImageBind-Huge  & 95.6    & 95.1 & 96.4 & 97.5    & 94.3   \\
		& DictAS  & ViT-L-14-336    & \textbf{98.2}    & \textbf{98.5} & \textbf{97.9} & \textbf{97.9}    & \textbf{96.4}   \\   \midrule
		\multirow{2}{*}{4-shot} & ResAD \cite{ResAD}   & ImageBind-Huge  & 96.9    & 97.5 & 96.8 & 97.9    & 96.1   \\
		& DictAS  & ViT-L-14-336    & \textbf{98.6}    & \textbf{98.8} & \textbf{98.0} & \textbf{98.4}    & \textbf{97.3}    \\ \bottomrule
		\end{tabular}
}
\end{table*}
 Table \ref{SOTA} compares our DictAS with two contemporary state-of-the-art methods, MetaAUS \cite{MetaUAS} and ResAD \cite{ResAD}. As our method currently applies only to transformer-based architectures, we selected the CLIP pre-trained backbones with the smallest (ViT-B-16) and largest (ViT-L-14-336) parameter counts for comparison. The experimental results show that, among the reported results, our DictAS achieves state-of-the-art performance in FSAS. Notably, despite using fewer backbone parameters than ResAD (which adopts ImageBird-Huge), our ViT-L-14-336-based DictAS performs better, highlighting its effectiveness.

\section{Additional Ablations}
\label{sec3}
\subsection{Ablation on Hyperparameters}
\begin{figure*}[tp]
	\centering
	\begin{subfigure}[b]{0.49\textwidth}
		\centering
		\includegraphics[width=\textwidth]{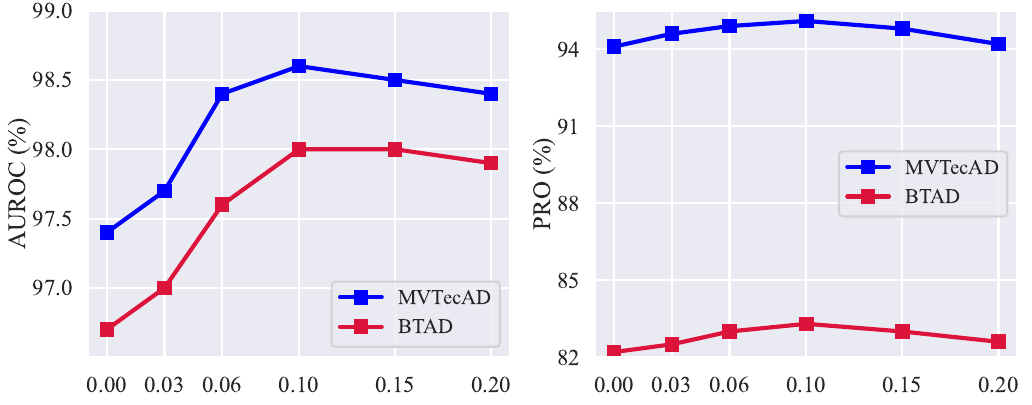}
		\caption{$\lambda_1$}
		\label{lambda1}
	\end{subfigure}
	\hfill
	\begin{subfigure}[b]{0.49\textwidth}
		\centering
		\includegraphics[width=\textwidth]{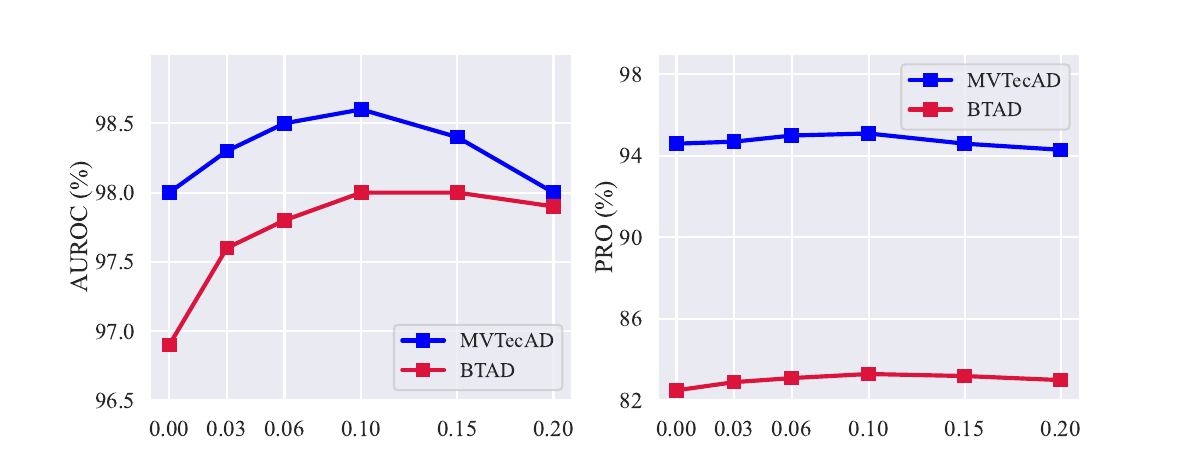}
		\caption{$\lambda_2$}
		\label{lambda2}
	\end{subfigure}
	\caption{(a) Ablation study on the weight coefficient $\lambda_1$ of the query contrastive constraint. (b) Ablation study on the weight coefficient $\lambda_2$ of the text alignment constraint. The experiments are conducted on the MVTecAD and BTAD datasets under the 4-shot setting and the metric pixel-level AUROC and PRO are reported.} 
	\label{lambda}
\end{figure*}
In this subsection, we conduct an ablation study on the weighting coefficients $\lambda_1$ and $\lambda_2$, which correspond to the two query discriminative regularization terms in our method. As shown in Figure \ref{lambda}, the model achieves optimal performance when both hyperparameters are set to approximately 0.1. As the weighting coefficients gradually increase to the equilibrium point (0.1), both AUROC and AP exhibit an upward trend. Beyond this point, the model's performance on unseen classes begins to gradually decline. 
\par 
\textbf{Reason Analysis.} Before analyzing the reasons, it is crucial to clarify the pseudo-labels used during the training process of DictAS under the self-supervised learning paradigm. The main loss, i.e., the query loss, is computed using all normal patches in the query image, where the query image feature itself serves as the pseudo-label. In contrast, the two query discriminative regularization terms use the synthesized mask $\mathbf{G}$ as the pseudo-label, which indicates the location of the synthesized anomaly within the query image.From this perspective, the query loss enables the model to acquire a category-agnostic dictionary querying capability, thereby facilitating generalization to unseen categories. Meanwhile, the two regularization losses leverage the synthetic anomaly information to enhance anomaly discriminability, making the boundary between normal and anomalous regions more distinguishable. Therefore, moderate regularization (e.g., $\lambda_1 = \lambda_2 = 0.1$) proves beneficial in the early training stages, as it improves the model’s ability to distinguish anomalies without overwhelming the dictionary querying mechanism.  However, as the influence of the regularization losses increases, the model's reliance on the dictionary-based querying diminishes. This shift causes the model to focus more on discriminating the synthesized anomalies during training, leading to a loss of generalization capability.

\subsection{Ablation on Auxiliary Datasets}

\begin{table}[t]
	\centering
	\begin{minipage}[t]{0.48\textwidth}
		\centering
		\caption{Ablation on different auxiliary datasets under 4-shot setting (\%). }
		\label{Aux_Data}
		\renewcommand{\arraystretch}{1.4}
		\resizebox{1.0\columnwidth}{!}
		{
			\begin{tabular}{>{\centering\arraybackslash}p{2.1cm}*{6}{>{\centering\arraybackslash}p{0.9cm}}}
				\toprule
				\multirow{2}{*}{\makecell[c]{Auxiliary \\ Dataset}} & \multicolumn{3}{c}{MVTecAD} & \multicolumn{3}{c}{BTAD} \\  \cmidrule(lr){2-4} \cmidrule(lr){5-7}
				& AUROC    & PRO     & AP     & AUROC   & PRO    & AP    \\   \midrule    
				VisA \cite{visa}                               & \textbf{98.6}     & \textbf{95.1}    & \textbf{66.8}   & 98.0    & 83.3   & 66.8  \\
				BrasTS \cite{BrasTS}                               & 98.3     & 95.0    & 66.2   & 97.9    & 83.0   & 66.2  \\
				Ade20K \cite{Ade20K}                              & 98.4     & \textbf{95.1}    & 66.5   & 98.1    & \textbf{83.5}   & 66.8  \\
				VOC2012 \cite{VOC}                             & 98.3     & 94.9    & 66.4   & \textbf{98.2}    & 83.4   & \textbf{66.9}  \\  \bottomrule
			\end{tabular}
		}
	\end{minipage}\hfill
	\begin{minipage}[t]{0.47\textwidth}
		\centering
		\caption{Ablation on the scale of auxiliary dataset VisA under 4-shot setting (\%).}
		\label{Scale}
		\resizebox{1\columnwidth}{!}
		{
			\begin{tabular}{>{\centering\arraybackslash}p{1cm}*{6}{>{\centering\arraybackslash}p{0.9cm}}}
				\toprule
				\multirow{2}{*}{Scale} & \multicolumn{3}{c}{MVTecAD} & \multicolumn{3}{c}{BTAD} \\    \cmidrule(lr){2-4} \cmidrule(lr){5-7}
				& AUROC    & PRO     & AP     & AUROC   & PRO    & AP    \\  \midrule
				15\%                   & 96.8     & 92.0    & 62.8   & 96.1    & 80.2   & 63.3  \\
				35\%                   & 97.3     & 92.9    & 63.5   & 96.6    & 81.7   & 63.8  \\
				55\%                   & 98.0     & 93.5    & 64.6   & 97.1    & 82.0   & 64.2  \\
				75\%                   & 98.3     & 94.6    & 66.0   & 97.6    & 82.9   & 66.5  \\
				95\%                   & 98.5     & 95.1    & 66.6   & 98.0    & 83.2   & 66.7  \\
				100\%                  & \textbf{98.6}     & \textbf{95.1}    & \textbf{66.8}   & \textbf{98.0}    & \textbf{83.3}   & \textbf{66.8}  \\  \bottomrule
			\end{tabular}
		}
	\end{minipage}
\end{table}

As mentioned above, our DictAS by default uses the industrial dataset VisA \cite{visa} as an auxiliary dataset for self-supervised training and then directly performs few-shot anomaly segmentation on unseen classes in other datasets. This setup is designed to follow the settings of existing methods for a fairer comparison \cite{VAND, AnomalyGPT, ResAD}. Can our method use a more general dataset for auxiliary training? If so, how does the scale of the auxiliary data affect the model's FSAS performance? We will address these two questions in the following discussion. 
\par  
\begin{table*}[t]
	\centering
	\caption{The details of different types of data transformations .}
	\label{AugType}
	\begin{tabular}{llc}
		\toprule
		\textbf{Type} & \textbf{Transformation} & \textbf{Parameters} \\
		\midrule
		\multirow{4}{*}{\textbf{Geo. Trans.}} 
		& RandomRotate90 & $p=1.0$ \\
		& Rotate & $30^\circ \sim 270^\circ, p=1.0$ \\
		& HorizontalFlip & $p=0.5$ \\
		& VerticalFlip & $p=0.5$ \\
		\midrule
		\multirow{2}{*}{\textbf{Color Trans.}} 
		& RandomBrightnessContrast & $p=0.5$ \\
		& HueSaturationValue & $hue=20, sat=30, val=20, p=0.5$ \\
		\midrule
		\multirow{2}{*}{\textbf{Noise Dist.}} 
		& GaussNoise & $var=10.0 \sim 50.0, p=0.5$ \\
		& MotionBlur & $blur\_limit=5, p=0.5$ \\
		\midrule
		\multirow{2}{*}{\textbf{Occl. Aug.}} 
		& GridDropout & $ratio=0.3, p=0.5$ \\
		& CoarseDropout & $max\_holes=8, max\_size=32\times32, p=0.5$ \\
		\bottomrule
	\end{tabular}
\end{table*}

\begin{table*}[t]
	\caption{Ablation on different types of data transformations.} 
	\centering
	\label{Ablation_datatype}
	\renewcommand{\arraystretch}{1.2}
	\resizebox{0.8\columnwidth}{!}
	{
		\begin{tabular}{*{4}{>{\centering\arraybackslash}p{1.8cm}}*{3}{>{\centering\arraybackslash}p{1.0cm}}}
			\toprule
			Geo.   Trans. & Color Trans. & Noise Dist. & Occl. Aug. & AUROC & PRO  & AP   \\   \midrule
			\ding{52}             &       \ding{56}       &     \ding{56}        &     \ding{56}       & 98.3  & 94.9 & 66.0 \\
			\ding{56}	& \ding{52}            &       \ding{56}      &     \ding{56}       & 98.2  & 94.7 & 64.7 \\
			\ding{56}	&        \ding{56}      & \ding{52}           &     \ding{56}       & 98.1  & 94.7 & 64.8 \\
			\ding{56}	&      \ding{56}        &       \ding{56}      & \ding{52}          & 98.2  & 94.8 & 64.9 \\
			\ding{56}	& \ding{52}            & \ding{52}           &    \ding{56}        & 97.9  & 94.1 & 64.0 \\
			\rowcolor{lightgreen}
			\ding{52}             &       \ding{56}       &     \ding{56}        & \ding{52}          & \textbf{98.6}  & \textbf{95.1} & \textbf{66.8} \\
			\ding{52}             & \ding{52}            &     \ding{56}        &     \ding{56}       & 98.2  & 94.7 & 64.6 \\
			\ding{52}             &       \ding{56}       & \ding{52}           &     \ding{56}       & 98.2  & 94.6 & 64.5 \\
			\ding{52}             & \ding{52}            & \ding{52}           & \ding{52}          & 98.3  & 94.8 & 65.5  \\  \bottomrule
		\end{tabular}
	}
\end{table*}
\textbf{Domain of Auxiliary Datasets.} In Table \ref{Aux_Data}, we investigate the impact of using auxiliary datasets from different domains for self-supervised training and evaluate their 4-shot performance on MVTecAD \cite{mvtec} and BTAD \cite{BTAD}. Specifically, the VisA dataset \cite{visa} from the industrial domain, the BrasTS dataset \cite{BrasTS} from the medical domain, and the Ade20K \cite{Ade20K} and VOC2012 \cite{VOC} datasets from natural scenes are used as auxiliary datasets. For the natural scene datasets Ade20K \cite{Ade20K} and VOC2012 \cite{VOC},  we randomly select samples identical to those in the VisA training set for auxiliary training. Since each natural image may contain multiple categories, we use \textit{object} to replace \textit{[class]} in the text prompts. Note that our auxiliary datasets do not require pixel-level annotations. Experimental results show that our DictAS is not sensitive to the auxiliary datasets and demonstrates strong robustness across industrial, medical, and natural scene domains. It is attributed to the use of the self-supervised learning paradigm, which demonstrates that DictAS has learned a generalizable dictionary lookup capability and successfully transferred this ability to the class-generalizable FSAS task.
\par
\textbf{Scale of Auxiliary Datasets.}  In Table \ref{Scale}, we evaluate the impact of the auxiliary dataset size on model performance. Specifically, (15\%, 35\%, 55\%, 75\%, 95\%) of the VisA training set samples are randomly selected for self-supervised training, and the FSAS performance on MVTecAD and BTAD is assessed under the 4-shot setting. The experimental results show that as the dataset size increases, the FSAS performance of the proposed DictAS also improves. Even when trained on only half or less of the auxiliary data, the proposed DictAS already achieves satisfactory results, highlighting the efficiency of our training strategy. Moreover, DictAS demonstrates promising potential with larger-scale training data, which will be explored in our future work.
\subsection{Ablation on Types of Data Transformations}

In this subsection, we conduct an ablation study on the types of data transformations used to generate reference images in the self-supervised training process. 
\par 
Specifically, we predefined four types of data transformations: Geometric Transformations (Geo. Trans.), Color Transformations (Color Trans.), Noise Disturbance (Noise Dist.), and Occlusion Augmentation (Occl. Aug). The details and hyperparameters of different types of data transformations are presented in Table \ref{AugType}. To investigate the impact of different transformation types on the experiment, we conducted an ablation study on these four types of transformations, as shown in Table \ref{Ablation_datatype}. 
It can be observed that when only a single transformation type is used, Geometric Transformations provide the greatest performance gain for FSAS, especially in terms of pixel-level AP (66.0\%). This is because applying geometric transformations to raw images, such as random rotation and random flipping, simulates the most significant variations among normal reference images in real-world anomaly segmentation. During training, self-supervised learning enables the model to capture the correspondence between query and reference images under geometric transformations, which helps DictAS enhance its robustness to different reference images. Furthermore, among different transformation combinations, the combination of Geometric Transformations and Occlusion Augmentation achieved the best results, with scores of 98.6\% in AUROC, 95.1\% in PRO and 66.8\% in AP. We attribute this to the occlusion simulating missing parts in real scenarios, further enhancing the robustness of the dictionary lookup.
\par 
Considering the model's performance, this work defaults to using Geometric Transformations and Occlusion Augmentation as the data transformation methods.

\section{Limitations} 
\label{sec4}
Our DictAS has demonstrated the state-of-the-art ZSAD performance in seven industrial and medical datasets. However, it still faces several limitations in practical applications: 1) Our method aims to learn the dictionary lookup ability of human inspectors when encountering unseen classes. While this enables generalization to novel categories, the dictionary lookup task imposes a limitation, requiring a few normal reference images to construct the dictionary, making it unsuitable for zero-shot tasks; 2) This work does not investigate the impact of larger-scale auxiliary datasets on the model's FSAS performance. However, ablation studies on the VisA dataset suggest that DictAS has the potential to leverage large-scale datasets (even at an internet scale) for self-supervised training, enabling continuous performance improvement. In the future, we will further enhance the FSAS capability of DictAS by incorporating human prior knowledge, while enabling zero-shot generalization. Moreover, larger-scale auxiliary data will be leveraged to enhance the dictionary lookup capability of DictAS.
 
\section{Detailed FSAS Results}
\label{sec5}
In this section, we present a detailed comparison of different SOTA methods under the 1-, 2-, and 4-shot settings. As mentioned in the main text, since DictAS primarily focuses on anomaly segmentation, pixel-level AUROC, PRO, and AP are used as the default evaluation metrics. As a complement, this section also reports image-level AUROC, F1-Max, and AP to assess the performance of few-shot anomaly classification. The classification score for each image is obtained following the same strategy as APRIL-GAN \cite{VAND}.
\clearpage
\subsection{Detailed few-shot anomaly classification results}
\begin{table*}[h]
	\caption{\textbf{Performance comparison of anomaly classification with other SOTA methods under the 1-shot setting}. The best results are highlighted in \textcolor{darkred}{red}, and the second-best results are marked in \textcolor{blue}{blue}. The symbol $\dag$ denotes methods based on CLIP, and (a,b,c) represents image-level (AUROC, F1-max, AP). To ensure a fair comparison, all methods use the same normal reference images, and all CLIP-based methods employ the same backbone (ViT-L-14-336) and input resolution ($336\times 336$).} 
	\centering
	\label{TabE1}
	\renewcommand{\arraystretch}{1.2}
	\resizebox{1\columnwidth}{!}
	{
		\begin{tabular}{>{\centering\arraybackslash}p{2.2cm}*{5}{>{\centering\arraybackslash}p{2.7cm}}>{\centering\arraybackslash}p{2.9cm}>{\centering\arraybackslash}p{2.6cm}>{\columncolor{lightgreen}\centering\arraybackslash}p{2.6cm}}
			\toprule
			Datasets & \makecell[c]{RegAD \cite{RegAD} \\ (ECCV 22)}    & \makecell[c]{AnomalyGPT \cite{AnomalyGPT}\\(AAAI 24)} & \makecell[c]{FastRecon \cite{fastrecon}\\(ICCV 23)} & \makecell[c]{$\dag$ FastRecon+ \cite{fastrecon}\\ (ICCV 23)} & \makecell[c]{$\dag$ WinCLIP \cite{winclip}\\ (CVPR 23)}  & \makecell[c]{$\dag$ APRIL-GAN \cite{VAND}\\(CVPR 23)} & \makecell[c]{$\dag$ PromptAD \cite{PromptAD}\\(CVPR 24)} & \makecell[c]{$\dag$ DictAS\\(Ours)}             \\   \midrule
			\multicolumn{9}{c}{Industrial Datasets\quad (AUROC, F1-Max, AP)}                                                                                                                                                \\   \midrule
			MVTecAD  & (73.3, 87.1, 87.2) & (92.8, \textcolor{blue}{94.3}, 96.1)   & (83.7, 90.9, 91.6)  & (92.0, 93.4, 95.6)   & (92.6, 92.0, 96.1) & (91.1, 90.9, 95.6)  & (\textcolor{blue}{93.0}, 93.7, \textcolor{blue}{96.6}) & (\textcolor{darkred}{96.1}, \textcolor{darkred}{94.4}, \textcolor{darkred}{98.3}) \\
			VisA     & (69.3, 76.2, 72.2) & (86.4, \textcolor{blue}{84.4}, 87.4)   & (80.1, 82.3, 83.1)  & (81.0, 81.4, 82.3)   & (84.8, 82.8, 87.0) & (\textcolor{blue}{87.1}, 83.1, \textcolor{blue}{90.5})  & (85.2, 83.3, 86.8) & (\textcolor{darkred}{89.5}, \textcolor{darkred}{85.9}, \textcolor{darkred}{91.0}) \\
			MVTec3D  & (54.0, 88.4, 81.7) & (76.0, 90.3, 91.9)   & (63.5, 89.6, 86.5)  & (72.8, \textcolor{blue}{90.8}, 89.9)   & (\textcolor{darkred}{79.8}, 90.3, \textcolor{blue}{93.1}) & (75.3, 89.8, 91.1)  & (71.2, 90.1, 89.8) & (\textcolor{blue}{78.6}, \textcolor{darkred}{91.1}, \textcolor{darkred}{93.4}) \\
			MPDD     & (47.9, 72.9, 61.5) & (72.4, 79.3, 75.9)   & (62.2, 77.5, 67.9)  & (76.5, 80.3, 76.1)   & (\textcolor{blue}{79.9}, 80.9, 82.5) & (75.1, 80.1, 80.8)  & (79.3, \textcolor{blue}{81.6}, \textcolor{darkred}{83.5}) & (\textcolor{darkred}{81.3}, \textcolor{darkred}{83.5}, \textcolor{blue}{82.6}) \\
			BTAD     & (84.4, 78.2, 80.5) & (93.6, 89.7, 94.6)   & (86.2, 77.7, 81.7)  & (\textcolor{blue}{93.7}, \textcolor{blue}{92.0}, \textcolor{blue}{95.0})   & (89.5, 81.8, 86.3) & (86.5, 84.0, 88.5)  & (93.4, 90.4, 94.4) & (\textcolor{darkred}{96.2}, \textcolor{darkred}{92.8}, \textcolor{darkred}{97.3}) \\
			\textbf{Average}  & (65.8, 80.5, 76.6) & (84.2, 87.6, 89.2)   & (75.1, 83.6, 82.1)  & (83.2, 87.6, 87.8)   & (\textcolor{blue}{85.3}, 85.5, 89.2) & (83.0, 85.6, 89.3)  & (84.4, \textcolor{blue}{87.8}, \textcolor{blue}{90.2}) & (\textcolor{darkred}{88.3}, \textcolor{darkred}{89.5}, \textcolor{darkred}{92.5}) \\  \midrule
			\multicolumn{9}{c}{Medical Datasets\quad (AUROC, F1-Max, AP)}                                                                                                                                                   \\   \midrule
			RESC     & (55.9, 60.4, 46.6) & (86.8, 76.2, 83.4)   & (76.8, 72.5, 59.4)  & (82.8, 71.3, 80.4)   & (57.4, 60.7, 48.1) & (77.3, 69.5, 69.7)  & (\textcolor{blue}{87.4}, \textcolor{blue}{78.2}, \textcolor{blue}{84.3}) & (\textcolor{darkred}{89.9}, \textcolor{darkred}{79.0}, \textcolor{darkred}{89.6}) \\
			BrasTS   & (58.4, 83.0, 73.2) & (73.1, 85.8, 82.0)   & (61.8, 84.3, 75.7)  & (76.2, 86.5, 85.1)   & (\textcolor{blue}{86.6}, 87.4, \textcolor{blue}{92.5}) & (\textcolor{darkred}{86.8}, \textcolor{darkred}{88.9}, 92.5)  & (81.7, 87.5, 88.5) & (85.8, \textcolor{blue}{88.0}, \textcolor{darkred}{92.9}) \\
			\textbf{Average}  & (57.2, 71.7, 59.9) & (79.9, 81.0, 82.7)   & (69.3, 78.4, 67.6)  & (79.5, 78.9, 82.7)   & (72.0, 74.0, 70.7) & (82.1, 79.2, 81.1)  & (\textcolor{blue}{84.6}, \textcolor{blue}{82.8}, \textcolor{blue}{86.4}) & (\textcolor{darkred}{87.8}, \textcolor{darkred}{83.5}, \textcolor{darkred}{91.2})   \\   \midrule
		\end{tabular}
	}
\end{table*}

\begin{table*}[h]
	\caption{\textbf{Performance comparison of anomaly classification with other SOTA methods under the 2-shot setting}. The best results are highlighted in \textcolor{darkred}{red}, and the second-best results are marked in \textcolor{blue}{blue}. The symbol $\dag$ denotes methods based on CLIP, and (a,b,c) represents image-level (AUROC, F1-max, AP). To ensure a fair comparison, all methods use the same normal reference images, and all CLIP-based methods employ the same backbone (ViT-L-14-336) and input resolution ($336\times 336$).} 
	\centering
	\label{TabE2}
	\renewcommand{\arraystretch}{1.2}
	\resizebox{1\columnwidth}{!}
	{
		\begin{tabular}{>{\centering\arraybackslash}p{2.2cm}*{5}{>{\centering\arraybackslash}p{2.7cm}}>{\centering\arraybackslash}p{2.9cm}>{\centering\arraybackslash}p{2.6cm}>{\columncolor{lightgreen}\centering\arraybackslash}p{2.6cm}}
			\toprule
			Datasets & \makecell[c]{RegAD \cite{RegAD} \\ (ECCV 22)}    & \makecell[c]{AnomalyGPT \cite{AnomalyGPT}\\(AAAI 24)} & \makecell[c]{FastRecon \cite{fastrecon}\\(ICCV 23)} & \makecell[c]{$\dag$ FastRecon+ \cite{fastrecon}\\ (ICCV 23)} & \makecell[c]{$\dag$ WinCLIP \cite{winclip}\\ (CVPR 23)}  & \makecell[c]{$\dag$ APRIL-GAN \cite{VAND}\\(CVPR 23)} & \makecell[c]{$\dag$ PromptAD \cite{PromptAD}\\(CVPR 24)} & \makecell[c]{$\dag$ DictAS\\(Ours)}             \\   \midrule
			\multicolumn{9}{c}{Industrial Datasets\quad (AUROC, F1-Max, AP)}                                                                                                                                                \\   \midrule
			MVTecAD  & (76.6, 88.8, 88.9) & (94.4, 95.0, 97.0)   & (88.9, 93.6, 94.7)  & (94.2, 94.5, 96.5)   & (93.8, 93.0, 96.6) & (90.1, 91.0, 95.5)  & (\textcolor{blue}{95.4}, \textcolor{blue}{95.1}, \textcolor{blue}{97.7}) & (\textcolor{darkred}{97.4}, \textcolor{darkred}{96.6}, \textcolor{darkred}{98.9}) \\
			VisA     & (70.4, 75.8, 73.6) & (\textcolor{blue}{87.2}, \textcolor{blue}{84.1}, 88.8)   & (84.6, 82.9, 86.7)  & (81.1, 81.8, 81.3)   & (83.5, 81.3, 85.9) & (86.6, 82.6, \textcolor{blue}{90.4})  & (85.1, 83.0, 87.0) & (\textcolor{darkred}{90.2}, \textcolor{darkred}{86.6}, \textcolor{darkred}{91.3}) \\
			MVTec3D  & (55.1, 88.5, 82.0) & (81.2, \textcolor{darkred}{91.5}, 94.2)   & (65.5, 89.6, 88.0)  & (76.9, \textcolor{blue}{91.2}, 92.2)   & (\textcolor{blue}{81.4}, 90.5, \textcolor{blue}{94.5}) & (75.8, 90.0, 91.5)  & (75.6, 90.8, 92.0) & (\textcolor{darkred}{82.4}, \textcolor{blue}{91.2}, \textcolor{darkred}{94.7}) \\
			MPDD     & (52.5, 73.6, 62.2) & (79.7, 82.1, 81.1)   & (67.0, 78.4, 70.6)  & (81.8, 83.0, 81.4)   & (81.5, 81.0, 83.3) & (75.1, 79.4, 80.2)  & (\textcolor{blue}{83.3}, \textcolor{blue}{83.6}, \textcolor{darkred}{88.2}) & (\textcolor{darkred}{84.9}, \textcolor{darkred}{86.4}, \textcolor{blue}{85.4}) \\
			BTAD     & (88.9, 89.2, 92.1) & (93.4, 89.9, \textcolor{blue}{95.0})   & (89.4, 83.2, 86.4)  & (\textcolor{blue}{93.8}, \textcolor{blue}{90.3}, 94.9)   & (90.7, 84.2, 87.6) & (86.1, 84.2, 88.5)  & (92.7, 89.0, 94.4) & (\textcolor{darkred}{95.6}, \textcolor{darkred}{92.3}, \textcolor{darkred}{96.6}) \\
			\textbf{Average}  & (68.7, 83.2, 79.8) & (\textcolor{blue}{87.2}, \textcolor{blue}{88.5}, 91.2)   & (79.1, 85.5, 85.3)  & (85.6, 88.2, 89.3)   & (86.2, 86.0, 89.6) & (82.7, 85.4, 89.2)  & (86.4, 88.3, \textcolor{blue}{91.9}) & (\textcolor{darkred}{90.1}, \textcolor{darkred}{90.6}, \textcolor{darkred}{93.4}) \\ \midrule
			\multicolumn{9}{c}{Medical Datasets\quad (AUROC, F1-Max, AP)}                                                                                                                                                   \\  \midrule
			RESC     & (59.4, 62.0, 48.0) & (87.8, 78.2, 83.5)   & (77.6, 72.7, 60.6)  & (87.6, 75.7, 84.9)   & (60.3, 61.0, 50.6) & (78.3, 70.8, 71.0)  & (\textcolor{blue}{89.2}, \textcolor{blue}{79.6}, \textcolor{blue}{85.9}) & (\textcolor{darkred}{91.6}, \textcolor{darkred}{80.6}, \textcolor{darkred}{90.9}) \\
			BrasTS   & (57.4, 83.5, 72.0) & (74.9, 86.6, 83.3)   & (65.4, 84.6, 77.4)  & (75.8, 87.2, 83.6)   & (\textcolor{blue}{87.0}, 88.0, \textcolor{darkred}{93.4}) & (\textcolor{darkred}{87.5}, \textcolor{darkred}{89.2}, \textcolor{blue}{93.0})  & (83.0, 88.1, 89.3) & (85.5, \textcolor{blue}{88.6}, 92.4) \\
			\textbf{Average}  & (58.4, 72.8, 60.0) & (81.3, 82.4, 83.4)   & (71.5, 78.7, 69.0)  & (81.7, 81.5, 84.3)   & (73.6, 74.5, 72.0) & (82.9, 80.0, 82.0)  & (\textcolor{blue}{86.1}, \textcolor{blue}{83.8}, \textcolor{blue}{87.6}) & (\textcolor{darkred}{88.6}, \textcolor{darkred}{84.6}, \textcolor{darkred}{91.7})  \\   \midrule
		\end{tabular}
	}
\end{table*}

\begin{table*}[h]
	\caption{\textbf{Performance comparison of anomaly classification with other SOTA methods under the 4-shot setting}. The best results are highlighted in \textcolor{darkred}{red}, and the second-best results are marked in \textcolor{blue}{blue}. The symbol $\dag$ denotes methods based on CLIP, and (a,b,c) represents image-level (AUROC, F1-max, AP). To ensure a fair comparison, all methods use the same normal reference images, and all CLIP-based methods employ the same backbone (ViT-L-14-336) and input resolution ($336\times 336$).} 
	\centering
	\label{TabE3}
	\renewcommand{\arraystretch}{1.2}
	\resizebox{1\columnwidth}{!}
	{
		\begin{tabular}{>{\centering\arraybackslash}p{2.2cm}*{5}{>{\centering\arraybackslash}p{2.7cm}}>{\centering\arraybackslash}p{2.9cm}>{\centering\arraybackslash}p{2.6cm}>{\columncolor{lightgreen}\centering\arraybackslash}p{2.6cm}}
			\toprule
			Datasets & \makecell[c]{RegAD \cite{RegAD} \\ (ECCV 22)}    & \makecell[c]{AnomalyGPT \cite{AnomalyGPT}\\(AAAI 24)} & \makecell[c]{FastRecon \cite{fastrecon}\\(ICCV 23)} & \makecell[c]{$\dag$ FastRecon+ \cite{fastrecon}\\ (ICCV 23)} & \makecell[c]{$\dag$ WinCLIP \cite{winclip}\\ (CVPR 23)}  & \makecell[c]{$\dag$ APRIL-GAN \cite{VAND}\\(CVPR 23)} & \makecell[c]{$\dag$ PromptAD \cite{PromptAD}\\(CVPR 24)} & \makecell[c]{$\dag$ DictAS\\(Ours)}             \\   \midrule
			\multicolumn{9}{c}{Industrial Datasets\quad (AUROC, F1-Max, AP)}                                                                                                                                                \\  \midrule
			MVTec-AD & (83.4, 89.8, 91.7) & (\textcolor{blue}{97.0}, \textcolor{blue}{95.9}, \textcolor{blue}{98.0})   & (94.2, 90.9, 90.4)  & (96.2, 95.3, 97.2)   & (95.5, 94.0, 97.3) & (91.0, 91.6, 95.9)  & (95.9, 95.2, 97.5) & (\textcolor{darkred}{98.8}, \textcolor{darkred}{98.2}, \textcolor{darkred}{99.5}) \\
			VisA     & (72.0, 77.1, 73.9) & (\textcolor{blue}{91.4}, \textcolor{blue}{87.2}, \textcolor{blue}{92.6})   & (68.5, 77.1, 72.6)  & (84.4, 82.6, 85.1)   & (85.7, 82.8, 87.8) & (87.2, 83.3, 91.1)  & (87.5, 83.9, 89.2) & (\textcolor{darkred}{92.3}, \textcolor{darkred}{88.5}, \textcolor{darkred}{93.6}) \\
			MVTec3D  & (57.7, 88.4, 84.1) & (\textcolor{blue}{83.4}, \textcolor{darkred}{91.6}, \textcolor{blue}{95.1})   & (57.9, 88.4, 83.7)  & (81.4, \textcolor{blue}{91.4}, 93.7)   & (81.3, 90.8, 94.3) & (76.4, 90.0, 91.8)  & (79.5, 91.1, 93.5) & (\textcolor{darkred}{84.5}, \textcolor{darkred}{91.6}, \textcolor{darkred}{95.3}) \\
			MPDD     & (61.1, 75.9, 66.9) & (85.9, \textcolor{darkred}{88.5}, 89.0)   & (79.8, 78.5, 75.7)  & (81.9, 82.8, 81.0)   & (84.0, 83.1, 86.1) & (76.5, 80.7, 81.6)  & (\textcolor{darkred}{88.0}, 87.6, \textcolor{darkred}{92.6}) & (\textcolor{blue}{87.3}, \textcolor{blue}{87.8}, \textcolor{blue}{89.2}) \\
			BTAD     & (91.3, 91.2, 94.5) & (93.5, 91.0, 95.9)   & (68.3, 79.1, 75.1)  & (\textcolor{blue}{94.4}, \textcolor{blue}{91.8}, \textcolor{blue}{96.2})   & (91.7, 84.3, 88.0) & (86.1, 83.9, 88.4)  & (92.6, 91.0, 94.4) & (\textcolor{darkred}{96.5}, \textcolor{darkred}{92.7}, \textcolor{darkred}{97.2}) \\
			\textbf{Average}  & (73.1, 84.5, 82.2) & (\textcolor{blue}{90.2}, \textcolor{blue}{90.8}, \textcolor{blue}{94.2})   & (73.7, 82.8, 79.5)  & (87.6, 88.8, 90.6)   & (87.6, 87.0, 90.7) & (83.4, 85.9, 89.8)  & (88.7, 89.8, 93.4) & (\textcolor{darkred}{91.9}, \textcolor{darkred}{91.8}, \textcolor{darkred}{94.9}) \\  \midrule
			\multicolumn{9}{c}{Medical Datasets\quad (AUROC, F1-Max, AP)}                                                                                                                                                   \\  \midrule
			RESC     & (64.2, 63.7, 51.0) & (88.5, 78.8, 85.4)   & (70.8, 65.6, 56.5)  & (87.5, 76.4, 84.6)   & (63.8, 62.6, 54.0) & (78.3, 70.7, 71.2)  & (\textcolor{blue}{90.2}, \textcolor{darkred}{81.0}, \textcolor{blue}{87.3}) & (\textcolor{darkred}{91.2}, \textcolor{blue}{79.6}, \textcolor{darkred}{90.6}) \\
			BrasTS   & (63.3, 83.9, 75.5) & (79.4, 86.2, 87.8)   & (54.9, 82.6, 72.4)  & (78.6, 87.6, 86.4)   & (87.0, 88.0, 93.4) & (\textcolor{blue}{88.0}, \textcolor{darkred}{89.1}, \textcolor{blue}{93.5})  & (86.4, \textcolor{blue}{88.2}, 92.4) & (\textcolor{darkred}{88.4}, 88.9, \textcolor{darkred}{94.3}) \\
			\textbf{Average}  & (63.7, 73.8, 63.2) & (83.9, 82.5, 86.6)   & (62.9, 74.1, 64.4)  & (83.0, 82.0, 85.5)   & (75.4, 75.3, 73.7) & (83.2, 79.9, 82.4)  & (\textcolor{blue}{88.3}, \textcolor{darkred}{84.6}, \textcolor{blue}{89.8}) & (\textcolor{darkred}{89.8}, \textcolor{blue}{84.3}, \textcolor{darkred}{92.5})  \\  \bottomrule
		\end{tabular}
	}
\end{table*}
\clearpage

\subsection{Detailed few-shot anomaly segmentation results}

\begin{table*}[h]
	\caption{\textbf{Performance comparison of anomaly segmentation with other SOTA methods under the 1-shot setting}. The best results are highlighted in \textcolor{darkred}{red}, and the second-best results are marked in \textcolor{blue}{blue}. The symbol $\dag$ denotes methods based on CLIP, and (a,b,c) represents pixel-level (AUROC, PRO, AP). To ensure a fair comparison, all methods use the same normal reference images, and all CLIP-based methods employ the same backbone (ViT-L-14-336) and input resolution ($336\times 336$).} 
	\centering
	\label{TabE4}
	\renewcommand{\arraystretch}{1}
	\resizebox{1\columnwidth}{!}
	{
		\begin{tabular}{>{\centering\arraybackslash}p{2.2cm}*{5}{>{\centering\arraybackslash}p{2.7cm}}>{\centering\arraybackslash}p{2.9cm}>{\centering\arraybackslash}p{2.6cm}>{\columncolor{lightgreen}\centering\arraybackslash}p{2.6cm}}
			\toprule
			Datasets & \makecell[c]{RegAD \cite{RegAD} \\ (ECCV 22)}    & \makecell[c]{AnomalyGPT \cite{AnomalyGPT}\\(AAAI 24)} & \makecell[c]{FastRecon \cite{fastrecon}\\(ICCV 23)} & \makecell[c]{$\dag$ FastRecon+ \cite{fastrecon}\\ (ICCV 23)} & \makecell[c]{$\dag$ WinCLIP \cite{winclip}\\ (CVPR 23)}  & \makecell[c]{$\dag$ APRIL-GAN \cite{VAND}\\(CVPR 23)} & \makecell[c]{$\dag$ PromptAD \cite{PromptAD}\\(CVPR 24)} & \makecell[c]{$\dag$ DictAS\\(Ours)}             \\   \midrule
			\multicolumn{9}{c}{Industrial Datasets\quad (AUROC, PRO, AP)}                                                                                                                                                \\   \midrule
		MVTecAD     & (92.3, 76.8, 36.1) & (\textcolor{blue}{95.3}, 89.0, 48.8)   & (93.9, 82.6, 48.2)  & (95.1, 90.8, 50.5)   & (91.6, 82.0, 35.5) & (91.2, 84.5, 43.8)  & (95.2, \textcolor{blue}{90.9}, \textcolor{blue}{53.3}) & (\textcolor{darkred}{97.7}, \textcolor{darkred}{92.5}, \textcolor{darkred}{61.1}) \\
		VisA        & (93.3, 68.7, 17.9) & (87.4, 65.3, 16.8)   & (96.5, 81.6, \textcolor{blue}{31.8})  & (96.1, 84.3, 26.0)   & (95.3, 85.2, 19.2) & (95.9, 87.0, 29.3)  & (\textcolor{blue}{97.2}, \textcolor{blue}{88.4}, 29.1) & (\textcolor{darkred}{98.0}, \textcolor{darkred}{89.6}, \textcolor{darkred}{32.7}) \\
		MVTec3D     & (95.4, 84.5, 8.4)  & (95.5, 84.3, 22.2)   & (95.4, 83.6, 15.9)  & (96.7, 89.3, 30.8)   & (96.2, 86.4, 22.8) & (96.1, 88.4, \textcolor{blue}{31.6})  & (\textcolor{blue}{97.0}, \textcolor{blue}{89.7}, 29.9) & (\textcolor{darkred}{97.5}, \textcolor{darkred}{92.1}, \textcolor{darkred}{34.4}) \\
		MPDD        & (93.2, 74.6, 8.4)  & (\textcolor{blue}{96.6}, 89.9, \textcolor{blue}{31.3})   & (95.5, 84.1, 20.9)  & (96.2, 90.1, 30.7)   & (95.6, 86.7, 23.6) & (94.9, 85.1, 28.3)  & (96.0, \textcolor{blue}{90.4}, 30.1) & (\textcolor{darkred}{97.4}, \textcolor{darkred}{92.8}, \textcolor{darkred}{33.3}) \\
		BTAD        & (95.6, 68.9, 33.1) & (95.7, 71.7, 49.9)   & (95.9, 67.9, 42.8)  & (\textcolor{blue}{96.8}, \textcolor{blue}{80.6}, 60.0)   & (88.9, 61.7, 26.0) & (93.0, 73.4, 50.4)  & (96.1, 79.4, \textcolor{blue}{61.3}) & (\textcolor{darkred}{97.6}, \textcolor{darkred}{82.1}, \textcolor{darkred}{64.6}) \\
		\textbf{Average}     & (94.0, 74.7, 20.8) & (94.1, 80.0, 33.8)   & (95.4, 80.0, 31.9)  & (96.2, 87.0, 39.6)   & (93.5, 80.4, 25.4) & (94.2, 83.7, 36.7)  & (\textcolor{blue}{96.3}, \textcolor{blue}{87.8}, \textcolor{blue}{40.8}) & (\textcolor{darkred}{97.6}, \textcolor{darkred}{89.8}, \textcolor{darkred}{45.2}) \\   \midrule
		\multicolumn{9}{c}{Medical Datasets \quad (AUROC, PRO, AP)}                                                                                                                                                      \\   \midrule
		RESC        & (84.6, 53.2, 14.6) & (86.0, 58.5, 27.4)   & (93.0, 76.5, 31.7)  & (96.0, 83.6, 66.5)   & (92.3, 73.3, 33.4) & (93.0, 74.9, 54.1)  & (\textcolor{blue}{96.4}, \textcolor{blue}{85.8}, \textcolor{blue}{68.2}) & (\textcolor{darkred}{97.2}, \textcolor{darkred}{88.8}, \textcolor{darkred}{72.4}) \\
		BrasTS      & (91.3, 62.6, 17.5) & (94.2, 69.9, 30.1)   & (93.4, 66.8, 25.8)  & (95.4, 71.4, 39.0)   & (93.1, 64.2, 33.2) & (90.9, 62.7, 38.7)  & (\textcolor{blue}{95.9}, \textcolor{darkred}{74.8}, \textcolor{blue}{46.0}) & (\textcolor{darkred}{96.5}, \textcolor{blue}{74.5}, \textcolor{darkred}{52.1}) \\
		\textbf{Average} & (88.0, 57.9, 16.0) & (90.1, 64.2, 28.8)   & (93.2, 71.6, 28.7)  & (95.7, 77.5, 52.8)   & (92.7, 68.7, 33.3) & (92.0, 68.8, 46.4)  & (\textcolor{blue}{96.2}, \textcolor{blue}{80.3}, \textcolor{blue}{57.1}) & (\textcolor{darkred}{96.9}, \textcolor{darkred}{81.6}, \textcolor{darkred}{62.3})  \\ \bottomrule
		\end{tabular}
}
\end{table*}

\begin{table*}[h]
	\caption{\textbf{Performance comparison of anomaly segmentation with other SOTA methods under the 2-shot setting}. The best results are highlighted in \textcolor{darkred}{red}, and the second-best results are marked in \textcolor{blue}{blue}. The symbol $\dag$ denotes methods based on CLIP, and (a,b,c) represents pixel-level (AUROC, PRO, AP). To ensure a fair comparison, all methods use the same normal reference images, and all CLIP-based methods employ the same backbone (ViT-L-14-336) and input resolution ($336\times 336$).} 
	\centering
	\label{TabE5}
	\renewcommand{\arraystretch}{1.2}
	\resizebox{1\columnwidth}{!}
	{
		\begin{tabular}{>{\centering\arraybackslash}p{2.2cm}*{5}{>{\centering\arraybackslash}p{2.7cm}}>{\centering\arraybackslash}p{2.9cm}>{\centering\arraybackslash}p{2.6cm}>{\columncolor{lightgreen}\centering\arraybackslash}p{2.6cm}}
			\toprule
			Datasets & \makecell[c]{RegAD \cite{RegAD} \\ (ECCV 22)}    & \makecell[c]{AnomalyGPT \cite{AnomalyGPT}\\(AAAI 24)} & \makecell[c]{FastRecon \cite{fastrecon}\\(ICCV 23)} & \makecell[c]{$\dag$ FastRecon+ \cite{fastrecon}\\ (ICCV 23)} & \makecell[c]{$\dag$ WinCLIP \cite{winclip}\\ (CVPR 23)}  & \makecell[c]{$\dag$ APRIL-GAN \cite{VAND}\\(CVPR 23)} & \makecell[c]{$\dag$ PromptAD \cite{PromptAD}\\(CVPR 24)} & \makecell[c]{$\dag$ DictAS\\(Ours)}             \\   \midrule
			\multicolumn{9}{c}{Industrial Datasets\quad (AUROC, PRO, AP)}                                                                                                                                                \\   \midrule
		MVTecAD  & (94.5, 82.7, 42.1) & (\textcolor{blue}{95.9}, 90.2, 50.7)   & (95.3, 85.8, 50.5)  & (95.5, \textcolor{blue}{91.5}, 51.9)   & (91.9, 82.7, 37.4) & (91.6, 85.5, 45.1)  & (95.6, \textcolor{blue}{91.5}, \textcolor{blue}{54.8}) & (\textcolor{darkred}{98.2}, \textcolor{darkred}{94.2}, \textcolor{darkred}{63.9}) \\
		VisA     & (94.3, 70.2, 21.6) & (87.7, 65.0, 19.7)   & (97.5, 83.9, \textcolor{blue}{37.5})  & (96.6, 85.2, 30.6)   & (95.7, 85.9, 23.6) & (96.1, 86.8, 30.1)  & (\textcolor{blue}{97.7}, \textcolor{blue}{89.4}, 34.4) & (\textcolor{darkred}{98.5}, \textcolor{darkred}{91.1}, \textcolor{darkred}{39.0}) \\
		MVTec3D  & (95.9, 86.2, 10.0) & (95.8, 85.5, 24.1)   & (95.8, 85.0, 16.9)  & (96.8, 90.4, \textcolor{blue}{35.5})   & (96.4, 87.0, 23.5) & (96.3, 88.8, 32.3)  & (\textcolor{blue}{97.2}, \textcolor{blue}{90.6}, 33.1) & (\textcolor{darkred}{97.9}, \textcolor{darkred}{93.4}, \textcolor{darkred}{38.8}) \\
		MPDD     & (94.0, 79.3, 13.1) & (\textcolor{blue}{97.3}, 91.8, 34.5)   & (96.8, 89.1, 26.2)  & (96.8, \textcolor{blue}{92.7}, \textcolor{blue}{35.7})   & (96.5, 89.4, 26.8) & (95.1, 86.6, 30.2)  & (96.8, 92.6, 34.5) & (\textcolor{darkred}{97.9}, \textcolor{darkred}{94.6}, \textcolor{darkred}{38.0}) \\
		BTAD     & (96.9, 74.1, 42.3) & (96.0, 72.4, 50.6)   & (96.4, 71.1, 45.1)  & (\textcolor{blue}{97.2}, \textcolor{blue}{80.5}, 61.6)   & (89.6, 63.4, 27.5) & (93.2, 73.2, 50.8)  & (96.4, 79.6, \textcolor{blue}{62.3}) & (\textcolor{darkred}{97.9}, \textcolor{darkred}{82.4}, \textcolor{darkred}{66.1}) \\
		\textbf{Average}  & (95.1, 78.5, 25.8) & (94.5, 81.0, 35.9)   & (96.4, 83.0, 35.2)  & (96.6, 88.1, 43.1)   & (94.0, 81.7, 27.8) & (94.5, 84.2, 37.7)  & (\textcolor{blue}{96.7}, \textcolor{blue}{88.7}, \textcolor{blue}{43.8}) & (\textcolor{darkred}{98.1}, \textcolor{darkred}{91.1}, \textcolor{darkred}{49.2}) \\ \midrule
		\multicolumn{9}{c}{Medical Datasets\quad(AUROC, PRO, AP)}                                                                                                                                                   \\  \midrule
		RESC     & (85.9, 54.5, 15.1) & (86.3, 59.0, 27.9)   & (93.5, 75.6, 32.9)  & (96.2, 84.7, 68.4)   & (92.7, 74.6, 35.7) & (93.4, 76.5, 56.0)  & (\textcolor{blue}{96.7}, \textcolor{blue}{86.6}, \textcolor{blue}{69.9}) & (\textcolor{darkred}{97.4}, \textcolor{darkred}{89.6}, \textcolor{darkred}{74.1}) \\
		BrasTS   & (92.7, 66.0, 20.6) & (94.1, 70.2, 29.7)   & (93.4, 67.3, 25.6)  & (95.2, 71.7, 35.3)   & (93.0, 63.6, 32.9) & (90.9, 63.1, 38.8)  & (\textcolor{blue}{95.8}, \textcolor{darkred}{75.2}, \textcolor{blue}{45.7}) & (\textcolor{darkred}{96.4}, \textcolor{blue}{73.8}, \textcolor{darkred}{53.8}) \\
		\textbf{Average}  & (89.3, 60.3, 17.9) & (90.2, 64.6, 28.8)   & (93.4, 71.4, 29.2)  & (95.7, 78.2, 51.9)   & (92.8, 69.1, 34.3) & (92.2, 69.8, 47.4)  & (\textcolor{blue}{96.3}, \textcolor{blue}{80.9}, \textcolor{blue}{57.8}) & (\textcolor{darkred}{96.9}, \textcolor{darkred}{81.7}, \textcolor{darkred}{62.0})  \\   \midrule
		\end{tabular}
}
\end{table*}

\begin{table*}[h]
	\caption{\textbf{Performance comparison of anomaly segmentation with other SOTA methods under the 4-shot setting}. The best results are highlighted in \textcolor{darkred}{red}, and the second-best results are marked in \textcolor{blue}{blue}. The symbol $\dag$ denotes methods based on CLIP, and (a,b,c) represents pixel-level (AUROC, PRO, AP). To ensure a fair comparison, all methods use the same normal reference images, and all CLIP-based methods employ the same backbone (ViT-L-14-336) and input resolution ($336\times 336$).} 
	\centering
	\label{TabE6}
	\renewcommand{\arraystretch}{1.2}
	\resizebox{1\columnwidth}{!}
	{
		\begin{tabular}{>{\centering\arraybackslash}p{2.2cm}*{5}{>{\centering\arraybackslash}p{2.7cm}}>{\centering\arraybackslash}p{2.9cm}>{\centering\arraybackslash}p{2.6cm}>{\columncolor{lightgreen}\centering\arraybackslash}p{2.6cm}}
			\toprule
			Datasets & \makecell[c]{RegAD \cite{RegAD} \\ (ECCV 22)}    & \makecell[c]{AnomalyGPT \cite{AnomalyGPT}\\(AAAI 24)} & \makecell[c]{FastRecon \cite{fastrecon}\\(ICCV 23)} & \makecell[c]{$\dag$ FastRecon+ \cite{fastrecon}\\ (ICCV 23)} & \makecell[c]{$\dag$ WinCLIP \cite{winclip}\\ (CVPR 23)}  & \makecell[c]{$\dag$ APRIL-GAN \cite{VAND}\\(CVPR 23)} & \makecell[c]{$\dag$ PromptAD \cite{PromptAD}\\(CVPR 24)} & \makecell[c]{$\dag$ DictAS\\(Ours)}             \\   \midrule
			\multicolumn{9}{c}{Industrial Datasets\quad (AUROC, PRO, AP)}                                                                                                                                                \\   \midrule
			MVTecAD \cite{mvtec} & (95.7, 86.0, 46.5) & (\textcolor{blue}{96.4}, 91.2, 52.9)   & (95.9, 79.9, 47.0)  & (96.3, 92.2, 53.9)   & (92.4, 83.8, 39.2) & (92.2, 86.6, 46.6)  & (96.0, \textcolor{blue}{92.4}, \textcolor{blue}{57.5}) & (\textcolor{darkred}{98.6}, \textcolor{darkred}{95.1}, \textcolor{darkred}{66.8}) \\
			VisA \cite{visa}     & (94.7, 72.8, 21.4) & (96.5, 65.4, 20.8)   & (96.0, 77.7, 31.1)  & (97.0, 86.2, 32.5)   & (96.0, 86.5, 25.7) & (96.2, 86.6, 30.6)  & (\textcolor{blue}{97.9}, \textcolor{blue}{89.5}, \textcolor{blue}{37.5}) & (\textcolor{darkred}{98.8}, \textcolor{darkred}{91.9}, \textcolor{darkred}{41.8}) \\
			MVTec3D \cite{mvtec3D}  & (96.9, 89.2, 13.3) & (96.6, 87.4, 27.8)   & (95.6, 83.6, 12.9)  & (97.1, 91.8, \textcolor{blue}{39.2})   & (96.6, 87.9, 24.0) & (96.4, 89.1, 33.1)  & (\textcolor{blue}{97.7}, \textcolor{blue}{92.1}, 36.9) & (\textcolor{darkred}{98.4}, \textcolor{darkred}{94.9}, \textcolor{darkred}{44.2}) \\
			MPDD \cite{MPDD}     & (94.9, 83.3, 16.4) & (\textcolor{blue}{97.7}, 93.2, \textcolor{blue}{40.8})   & (97.0, 87.5, 25.7)  & (97.4, 93.1, 37.8)   & (97.0, 90.7, 29.3) & (95.3, 86.9, 31.4)  & (97.3, \textcolor{blue}{94.0}, 40.5) & (\textcolor{darkred}{98.4}, \textcolor{darkred}{95.8}, \textcolor{darkred}{42.9}) \\
			BTAD \cite{BTAD}    & (97.3, 75.5, 44.1) & (96.2, 73.5, 50.6)   & (88.7, 62.1, 35.5)  & (\textcolor{blue}{97.4}, \textcolor{blue}{80.8}, 62.2)   & (90.3, 64.7, 28.5) & (93.3, 74.6, 50.9)  & (96.6, 80.1, \textcolor{blue}{62.5}) & (\textcolor{darkred}{98.0}, \textcolor{darkred}{83.3}, \textcolor{darkred}{66.8}) \\
			\textbf{Average}  & (95.9, 81.3, 28.3) & (96.7, 82.1, 38.6)   & (94.6, 78.2, 30.4)  & (97.0, 88.8, 45.1)   & (94.5, 82.7, 29.3) & (94.7, 84.8, 38.5)  & (\textcolor{blue}{97.1}, \textcolor{blue}{89.6}, \textcolor{blue}{47.0}) & (\textcolor{darkred}{98.4}, \textcolor{darkred}{92.2}, \textcolor{darkred}{52.5}) \\   \midrule
			\multicolumn{9}{c}{Medical Datasets \quad (AUROC, PRO, AP)}                                                                                                                                                   \\   \midrule
			RESC \cite{RESC}   & (87.9, 60.0, 18.1) & (86.7, 60.0, 28.5)   & (91.7, 71.7, 30.3)  & (95.8, 82.8, 68.5)   & (93.1, 75.7, 38.4) & (93.7, 77.6, 57.3)  & (\textcolor{blue}{96.8}, \textcolor{blue}{86.8}, \textcolor{blue}{71.3}) & (\textcolor{darkred}{97.5}, \textcolor{darkred}{89.7}, \textcolor{darkred}{74.9}) \\
			BrasTS \cite{BrasTS}   & (93.8, 70.2, 24.8) & (95.4, 73.6, 41.8)   & (92.5, 63.8, 24.0)  & (96.1, 73.8, 43.9)   & (93.1, 64.0, 33.4) & (91.3, 63.0, 40.0)  & (\textcolor{blue}{96.6}, \textcolor{blue}{77.0}, \textcolor{blue}{54.4}) & (\textcolor{darkred}{97.3}, \textcolor{darkred}{77.2}, \textcolor{darkred}{59.3}) \\
			\textbf{Average}  & (90.8, 65.1, 21.5) & (91.0, 66.8, 35.2)   & (92.1, 67.8, 27.1)  & (96.0, 78.3, 56.2)   & (93.1, 69.8, 35.9) & (92.5, 70.3, 48.7)  & (\textcolor{blue}{96.7}, \textcolor{blue}{82.2}, \textcolor{blue}{62.9}) & (\textcolor{darkred}{97.4}, \textcolor{darkred}{83.4}, \textcolor{darkred}{67.1})  \\ \bottomrule
		\end{tabular}
	}
\end{table*}

\begin{table*}[!ht]
\caption{Anomaly segmentation performance of our DictAS on \textbf{MVTecAD} for each object category. Pixel-level AUROC, PRO and AP are reported.}
\centering
\label{TabE7}
\renewcommand{\arraystretch}{1.0}
	\begin{adjustbox}{width=1\columnwidth}
	\begin{tabular}{cccccccccc}
		\toprule
		\multirow{2}{*}{Object} & \multicolumn{3}{c}{1-shot}                    & \multicolumn{3}{c}{2-shot}                    & \multicolumn{3}{c}{4-shot}                    \\   \cmidrule(lr){2-4}   \cmidrule(lr){5-7}  \cmidrule(lr){8-10}
		& AUROC         & PRO         & AP            & AUROC         & PRO         & AP            & AUROC         & PRO         & AP            \\   \midrule
		bottle                  & 99.1\dev{0.1} & 96.7\dev{0.3} & 87.2\dev{0.9} & 99.2\dev{0.0} & 96.7\dev{0.3} & 87.5\dev{0.5} & 99.2\dev{0.0} & 96.6\dev{0.1} & 87.2\dev{0.6} \\
		cable                   & 97.8\dev{0.3} & 90.1\dev{0.6} & 66.7\dev{2.5} & 98.7\dev{0.4} & 93.7\dev{1.1} & 76.5\dev{4.2} & 98.9\dev{0.3} & 94.8\dev{0.9} & 78.7\dev{2.1} \\
		capsule                 & 97.7\dev{0.2} & 93.1\dev{0.9} & 37.3\dev{9.4} & 98.5\dev{0.3} & 95.6\dev{1.2} & 43.0\dev{9.2} & 98.6\dev{0.3} & 95.6\dev{0.8} & 45.0\dev{4.4} \\
		carpet                  & 99.7\dev{0.0} & 98.4\dev{0.1} & 85.1\dev{0.3} & 99.7\dev{0.0} & 98.5\dev{0.1} & 85.1\dev{0.3} & 99.7\dev{0.0} & 98.4\dev{0.0} & 85.4\dev{0.3} \\
		grid                    & 96.3\dev{0.7} & 88.4\dev{2.1} & 33.2\dev{0.5} & 96.9\dev{0.6} & 90.2\dev{1.6} & 33.0\dev{2.3} & 97.7\dev{0.6} & 92.9\dev{2.2} & 36.4\dev{1.0} \\
		hazelnut                & 98.4\dev{0.2} & 94.2\dev{1.2} & 62.0\dev{1.8} & 98.8\dev{0.3} & 95.5\dev{0.8} & 64.8\dev{2.5} & 99.1\dev{0.1} & 96.2\dev{0.2} & 67.0\dev{1.7} \\
		leather                 & 99.6\dev{0.0} & 98.8\dev{0.1} & 57.9\dev{0.4} & 99.6\dev{0.0} & 98.8\dev{0.1} & 57.5\dev{0.5} & 99.6\dev{0.0} & 98.7\dev{0.1} & 58.5\dev{1.0} \\
		metal\_nut               & 95.8\dev{0.9} & 93.3\dev{1.0} & 74.7\dev{4.2} & 96.0\dev{0.6} & 94.3\dev{1.3} & 75.3\dev{3.2} & 97.5\dev{0.1} & 96.3\dev{0.3} & 82.5\dev{1.0} \\
		pill                    & 98.5\dev{0.1} & 97.8\dev{0.1} & 77.4\dev{0.8} & 98.7\dev{0.1} & 97.9\dev{0.1} & 80.1\dev{0.9} & 98.9\dev{0.1} & 98.0\dev{0.1} & 81.8\dev{0.9} \\
		screw                   & 98.3\dev{0.6} & 92.0\dev{1.7} & 30.7\dev{0.7} & 98.6\dev{0.7} & 93.5\dev{2.4} & 26.2\dev{0.2} & 99.1\dev{0.8} & 94.7\dev{3.1} & 37.8\dev{1.0} \\
		tile                    & 98.5\dev{0.1} & 95.8\dev{0.3} & 82.1\dev{1.2} & 98.6\dev{0.1} & 96.1\dev{0.2} & 83.0\dev{0.6} & 98.8\dev{0.0} & 96.3\dev{0.2} & 85.0\dev{0.1} \\
		toothbrush              & 97.3\dev{0.8} & 85.4\dev{3.1} & 40.2\dev{6.0} & 99.0\dev{0.6} & 91.2\dev{2.5} & 52.7\dev{4.3} & 99.2\dev{0.4} & 91.4\dev{4.2} & 56.7\dev{4.5} \\
		transistor              & 93.3\dev{2.1} & 75.9\dev{4.1} & 56.1\dev{5.7} & 95.7\dev{1.2} & 82.8\dev{3.5} & 63.5\dev{4.4} & 96.5\dev{0.7} & 87.2\dev{1.9} & 66.1\dev{3.0} \\
		wood                    & 97.1\dev{0.1} & 94.5\dev{0.2} & 70.9\dev{0.3} & 97.3\dev{0.1} & 94.5\dev{0.1} & 71.8\dev{0.3} & 97.4\dev{0.1} & 94.5\dev{0.2} & 72.5\dev{0.7} \\
		zipper                  & 97.8\dev{0.0} & 93.9\dev{0.1} & 55.2\dev{0.3} & 98.0\dev{0.2} & 94.4\dev{0.5} & 58.0\dev{0.7} & 98.3\dev{0.1} & 95.0\dev{0.3} & 60.8\dev{1.2} \\   \midrule
		\textbf{Average}                    & 97.7\dev{0.1} & 92.5\dev{0.3} & 61.1\dev{0.5} & 98.2\dev{0.1} & 94.2\dev{0.2} & 63.9\dev{1.2} & 98.6\dev{0.0} & 95.1\dev{0.3} & 66.8\dev{0.4}  \\  \bottomrule
	\end{tabular}
\end{adjustbox}
\end{table*}

\begin{table*}[!ht]
	\caption{Anomaly segmentation performance of our DictAS on \textbf{VisA} for each object category. Pixel-level AUROC, PRO and AP are reported.}
	\centering
	\label{TabE8}
	\renewcommand{\arraystretch}{1.0}
	\begin{adjustbox}{width=1\columnwidth}
	\begin{tabular}{cccccccccc}
		\toprule
		\multirow{2}{*}{Object} & \multicolumn{3}{c}{1-shot}                    & \multicolumn{3}{c}{2-shot}                     & \multicolumn{3}{c}{4-shot}                     \\   \cmidrule(lr){2-4}   \cmidrule(lr){5-7}  \cmidrule(lr){8-10}
	& AUROC         & PRO         & AP            & AUROC         & PRO         & AP            & AUROC         & PRO         & AP            \\  \midrule
		candle                  & 99.3\dev{0.1} & 96.3\dev{0.1} & 23.6\dev{0.9} & 99.4\dev{0.1} & 96.4\dev{0.1} & 23.7\dev{0.5}  & 99.5\dev{0.0} & 96.7\dev{0.1} & 24.5\dev{0.6}  \\
		capsules                & 97.8\dev{0.2} & 84.6\dev{2.3} & 37.0\dev{1.2} & 98.4\dev{0.2} & 86.0\dev{1.4} & 39.7\dev{1.1}  & 98.7\dev{0.1} & 87.6\dev{2.1} & 40.0\dev{0.7}  \\
		cashew                  & 99.4\dev{0.1} & 96.1\dev{0.6} & 60.3\dev{2.8} & 99.5\dev{0.1} & 96.2\dev{0.4} & 66.4\dev{2.8}  & 99.5\dev{0.0} & 95.7\dev{0.3} & 67.5\dev{2.1}  \\
		chewinggum              & 99.6\dev{0.0} & 93.0\dev{0.4} & 78.1\dev{0.5} & 99.6\dev{0.0} & 91.8\dev{0.6} & 78.8\dev{0.5}  & 99.6\dev{0.0} & 92.4\dev{0.3} & 78.1\dev{0.3}  \\
		fryum                   & 97.5\dev{0.2} & 89.5\dev{0.5} & 41.5\dev{1.1} & 97.8\dev{0.2} & 90.1\dev{0.8} & 42.9\dev{1.4}  & 97.9\dev{0.1} & 90.8\dev{0.8} & 44.0\dev{0.3}  \\
		macaroni1               & 99.2\dev{0.2} & 96.5\dev{1.7} & 10.4\dev{0.7} & 99.5\dev{0.1} & 97.5\dev{0.5} & 12.1\dev{0.7}  & 99.6\dev{0.0} & 97.6\dev{0.2} & 15.1\dev{0.4}  \\
		macaroni2               & 96.7\dev{0.6} & 86.7\dev{1.8} & 2.7\dev{1.3}  & 96.6\dev{0.5} & 87.5\dev{0.4} & 5.4\dev{0.8}   & 97.5\dev{0.3} & 91.0\dev{1.2} & 7.1\dev{0.8}   \\
		pcb1                    & 98.4\dev{0.5} & 91.3\dev{3.7} & 43.3\dev{4.6} & 99.4\dev{0.1} & 93.7\dev{1.8} & 73.3\dev{4.0} & 99.6\dev{0.1} & 94.9\dev{1.3} & 81.1\dev{3.8} \\
		pcb2                    & 96.2\dev{0.3} & 76.0\dev{3.4} & 12.6\dev{3.7} & 97.2\dev{0.2} & 81.8\dev{2.6} & 19.1\dev{2.7}  & 97.5\dev{0.2} & 80.1\dev{1.7} & 20.5\dev{1.9}  \\
		pcb3                    & 95.4\dev{0.4} & 79.5\dev{3.0} & 13.6\dev{1.4} & 97.1\dev{0.4} & 85.7\dev{2.0} & 25.9\dev{1.4}  & 97.9\dev{0.1} & 87.8\dev{2.0} & 30.8\dev{1.1}  \\
		pcb4                    & 97.5\dev{0.4} & 89.0\dev{2.5} & 18.4\dev{3.7} & 98.1\dev{0.2} & 89.8\dev{0.6} & 29.9\dev{6.3}  & 98.6\dev{0.3} & 91.8\dev{1.0} & 40.6\dev{8.8}  \\
		pipe\_fryum              & 99.1\dev{0.1} & 97.0\dev{0.3} & 50.8\dev{1.6} & 99.2\dev{0.1} & 96.8\dev{0.2} & 50.4\dev{2.5}  & 99.2\dev{0.0} & 96.8\dev{0.2} & 51.9\dev{0.7}  \\  \midrule
		\textbf{Average}                    & 98.0\dev{0.1} & 89.6\dev{0.7} & 32.7\dev{0.9} & 98.5\dev{0.1} & 91.1\dev{0.4} & 39.0\dev{2.0}  & 98.8\dev{0.1} & 91.9\dev{0.3} & 41.8\dev{1.7} \\  \bottomrule
	\end{tabular}
\end{adjustbox}
\end{table*}

\begin{table*}[!ht]
	\caption{Anomaly segmentation performance of our DictAS on \textbf{MVTec3D} for each object category. Pixel-level AUROC, PRO and AP are reported.}
	\centering
	\label{TabE9}
	\renewcommand{\arraystretch}{1.0}
	\begin{adjustbox}{width=1\columnwidth}
		\begin{tabular}{cccccccccc}
			\toprule
		\multirow{2}{*}{Object} & \multicolumn{3}{c}{1-shot}                     & \multicolumn{3}{c}{2-shot}                     & \multicolumn{3}{c}{4-shot}                    \\
			\cmidrule(lr){2-4}   \cmidrule(lr){5-7}  \cmidrule(lr){8-10}
	& AUROC         & PRO         & AP            & AUROC         & PRO         & AP            & AUROC         & PRO         & AP           \\  \midrule
		cookie                  & 98.7\dev{0.1} & 94.5\dev{0.2} & 60.3\dev{2.8}  & 98.9\dev{0.1} & 95.2\dev{0.3} & 65.0\dev{1.3}  & 99.1\dev{0.1} & 96.1\dev{0.4} & 69.1\dev{1.1} \\
		dowel                   & 98.1\dev{0.2} & 92.4\dev{0.7} & 24.1\dev{2.9}  & 98.4\dev{0.3} & 94.0\dev{1.2} & 24.5\dev{3.0}  & 99.1\dev{0.3} & 96.0\dev{1.2} & 32.7\dev{6.1} \\
		cable\_gland             & 96.1\dev{0.7} & 88.1\dev{1.7} & 11.8\dev{2.2}  & 97.2\dev{0.5} & 91.6\dev{1.5} & 16.4\dev{5.3}  & 99.1\dev{0.6} & 97.4\dev{1.5} & 32.6\dev{3.7} \\
		rope                    & 99.2\dev{0.1} & 96.7\dev{0.3} & 45.8\dev{1.6}  & 99.2\dev{0.1} & 96.7\dev{0.3} & 44.3\dev{1.2}  & 99.3\dev{0.0} & 97.4\dev{0.1} & 46.8\dev{0.8} \\
		peach                   & 98.5\dev{0.5} & 94.5\dev{1.8} & 26.6\dev{15.5} & 99.1\dev{0.5} & 96.6\dev{1.8} & 39.5\dev{16.6} & 99.6\dev{0.0} & 98.5\dev{0.2} & 56.0\dev{1.7} \\
		potato                  & 99.4\dev{0.1} & 97.3\dev{0.2} & 29.8\dev{2.1}  & 99.3\dev{0.0} & 97.0\dev{0.2} & 30.3\dev{1.9}  & 99.5\dev{0.1} & 97.7\dev{0.3} & 35.2\dev{2.5} \\
		bagel                   & 99.5\dev{0.0} & 98.3\dev{0.2} & 67.1\dev{1.7}  & 99.5\dev{0.0} & 98.4\dev{0.2} & 66.5\dev{0.9}  & 99.6\dev{0.0} & 98.6\dev{0.3} & 65.7\dev{1.8} \\
		carrot                  & 99.4\dev{0.1} & 97.7\dev{0.2} & 31.3\dev{1.0}  & 99.4\dev{0.1} & 98.0\dev{0.2} & 34.4\dev{2.0}  & 99.5\dev{0.0} & 98.2\dev{0.2} & 35.1\dev{1.4} \\
		foam                    & 88.6\dev{1.0} & 69.6\dev{2.1} & 30.9\dev{0.6}  & 88.8\dev{0.4} & 70.8\dev{1.1} & 31.0\dev{0.3}  & 90.0\dev{0.2} & 72.1\dev{0.6} & 30.9\dev{0.1} \\
		tire                    & 98.0\dev{0.1} & 91.5\dev{0.3} & 16.2\dev{0.8}  & 99.1\dev{0.1} & 95.6\dev{0.4} & 35.6\dev{0.9}  & 99.3\dev{0.0} & 96.5\dev{0.3} & 37.5\dev{0.8} \\  \midrule
		\textbf{Average}                    & 97.5\dev{0.1} & 92.1\dev{0.1} & 34.4\dev{1.5}  & 97.9\dev{0.1} & 93.4\dev{0.3} & 38.8\dev{1.7}  & 98.4\dev{0.1} & 94.9\dev{0.2} & 44.2\dev{1.1} \\
    \bottomrule
\end{tabular}
\end{adjustbox}
\end{table*}

\begin{table*}[!ht]
	\caption{Anomaly segmentation performance of our DictAS on \textbf{MPDD} for each object category. Pixel-level AUROC, PRO and AP are reported.}
	\centering
	\label{TabE10}
	\renewcommand{\arraystretch}{1.0}
	\begin{adjustbox}{width=1\columnwidth}
		\begin{tabular}{cccccccccc}
			\toprule
		\multirow{2}{*}{Object} & \multicolumn{3}{c}{1-shot}                    & \multicolumn{3}{c}{2-shot}                    & \multicolumn{3}{c}{4-shot}                     \\
			\cmidrule(lr){2-4}   \cmidrule(lr){5-7}  \cmidrule(lr){8-10}
	& AUROC         & PRO         & AP            & AUROC         & PRO         & AP            & AUROC         & PRO         & AP           \\ \midrule
		bracket\_brown           & 94.8\dev{0.4} & 90.7\dev{0.7} & 5.3\dev{0.3}  & 95.7\dev{0.3} & 92.9\dev{1.3} & 7.1\dev{0.7}  & 96.5\dev{0.3} & 94.3\dev{1.1} & 9.7\dev{0.9}   \\
		connector               & 97.3\dev{0.4} & 90.9\dev{1.4} & 23.0\dev{3.9} & 97.8\dev{0.2} & 92.3\dev{0.8} & 28.8\dev{3.7} & 98.5\dev{0.2} & 94.8\dev{0.7} & 51.2\dev{3.3} \\
		tubes                   & 99.2\dev{0.1} & 97.0\dev{0.3} & 73.0\dev{1.6} & 99.4\dev{0.1} & 97.8\dev{0.3} & 75.5\dev{1.1} & 99.5\dev{0.1} & 98.2\dev{0.3} & 75.8\dev{0.8}  \\
		metal\_plate             & 98.3\dev{0.0} & 95.0\dev{0.1} & 89.8\dev{0.1} & 99.0\dev{0.0} & 96.4\dev{0.1} & 93.3\dev{0.2} & 99.2\dev{0.1} & 96.8\dev{0.2} & 94.4\dev{0.4}  \\
		bracket\_black           & 95.0\dev{1.5} & 89.6\dev{5.7} & 4.0\dev{3.6}  & 95.7\dev{0.7} & 91.9\dev{2.0} & 11.8\dev{0.9} & 96.7\dev{1.4} & 93.7\dev{4.2} & 13.4\dev{5.5}  \\
		bracket\_white           & 99.4\dev{0.1} & 93.4\dev{3.1} & 4.8\dev{0.6}  & 99.8\dev{0.1} & 96.3\dev{1.9} & 11.8\dev{0.4} & 99.8\dev{0.2} & 97.0\dev{0.6} & 12.7\dev{2.5}  \\  \midrule
		\textbf{Average}                    & 97.4\dev{0.2} & 92.8\dev{0.5} & 33.3\dev{0.2} & 97.9\dev{0.1} & 94.6\dev{0.3} & 38.0\dev{1.7} & 98.4\dev{0.3} & 95.8\dev{0.8} & 42.9\dev{1.7}  \\
    \bottomrule
\end{tabular}
\end{adjustbox}
\end{table*}

\begin{table*}[!ht]
	\caption{Anomaly segmentation performance of our DictAS on \textbf{BTAD} for each object category. Pixel-level AUROC, PRO and AP are reported.}
	\centering
	\label{TabE11}
	\renewcommand{\arraystretch}{1.0}
	\begin{adjustbox}{width=1\columnwidth}
		\begin{tabular}{cccccccccc}
			\toprule
		\multirow{2}{*}{Object} & \multicolumn{3}{c}{1-shot}                    & \multicolumn{3}{c}{2-shot}                    & \multicolumn{3}{c}{4-shot}                    \\
			\cmidrule(lr){2-4}   \cmidrule(lr){5-7}  \cmidrule(lr){8-10}
	& AUROC         & PRO         & AP            & AUROC         & PRO         & AP            & AUROC         & PRO         & AP           \\  \midrule
		01                  & 97.0\dev{0.2} & 77.2\dev{1.5} & 60.5\dev{0.9} & 97.3\dev{0.1} & 78.9\dev{0.5} & 61.4\dev{0.5} & 97.5\dev{0.1} & 80.6\dev{0.6} & 61.8\dev{0.4} \\   
		02                    & 97.0\dev{0.0} & 73.1\dev{1.5} & 74.2\dev{0.4} & 97.1\dev{0.1} & 71.5\dev{1.3} & 74.5\dev{0.8} & 97.2\dev{0.0} & 72.2\dev{0.6} & 74.4\dev{0.3} \\
		03                  & 99.0\dev{0.1} & 96.0\dev{0.1} & 59.1\dev{1.4} & 99.1\dev{0.1} & 96.6\dev{0.3} & 62.5\dev{1.6} & 99.3\dev{0.0} & 97.2\dev{0.1} & 64.1\dev{1.8} \\
		\midrule
		\textbf{Average}                    & 97.6\dev{0.1} & 82.1\dev{0.6} & 64.6\dev{0.7} & 97.9\dev{0.1} & 82.4\dev{0.3} & 66.1\dev{0.8} & 98.0\dev{0.0} & 83.3\dev{0.4} & 66.8\dev{0.7}  \\ 
    \bottomrule
\end{tabular}
\end{adjustbox}
\end{table*}

\clearpage

\section{Detailed Qualitative Results}

\begin{figure*}[h]
	\centering
	\includegraphics[width=0.9\columnwidth]{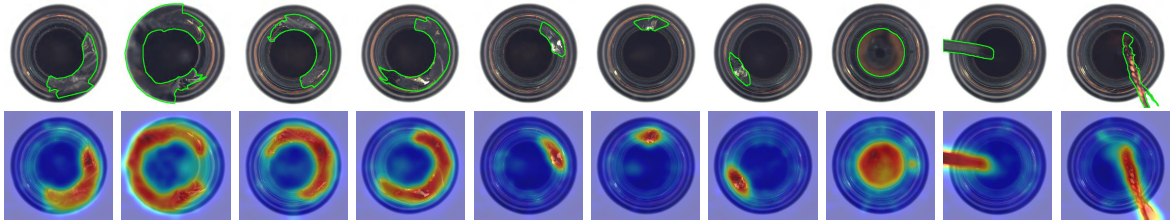}
	\caption{Visualization of segmentation results for the \textbf{bottle} class on \textbf{MVTecAD} under the 4-shot setting. The first row displays the input images, with green outlines indicating the ground truth regions. The second row presents the anomaly segmentation results.}
	\label{fig9}
\end{figure*}

\begin{figure*}[h]
	\centering
	\includegraphics[width=0.9\columnwidth]{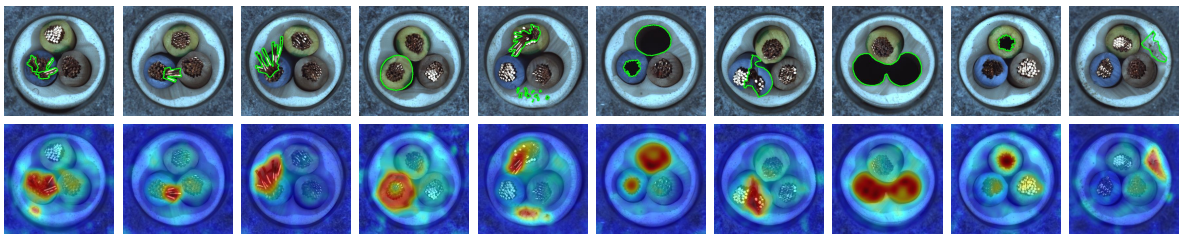}
	\caption{Visualization of segmentation results for the \textbf{cable} class on \textbf{MVTecAD} under the 4-shot setting. The first row displays the input images, with green outlines indicating the ground truth regions. The second row presents the anomaly segmentation results.}
	\label{fig10}
\end{figure*}

\begin{figure*}[h]
	\centering
	\includegraphics[width=0.9\columnwidth]{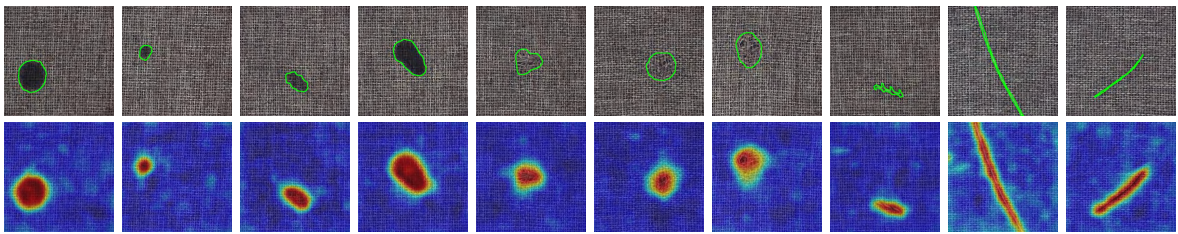}
	\caption{Visualization of segmentation results for the \textbf{carpet} class on \textbf{MVTecAD} under the 4-shot setting. The first row displays the input images, with green outlines indicating the ground truth regions. The second row presents the anomaly segmentation results.}
	\label{fig11}
\end{figure*}

\begin{figure*}[h]
	\centering
	\includegraphics[width=0.9\columnwidth]{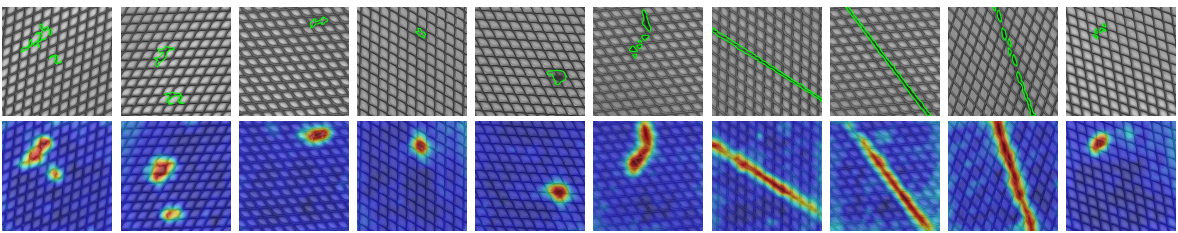}
	\caption{Visualization of segmentation results for the \textbf{grid} class on \textbf{MVTecAD} under the 4-shot setting. The first row displays the input images, with green outlines indicating the ground truth regions. The second row presents the anomaly segmentation results.}
	\label{fig12}
\end{figure*}

\begin{figure*}[h]
	\centering
	\includegraphics[width=0.9\columnwidth]{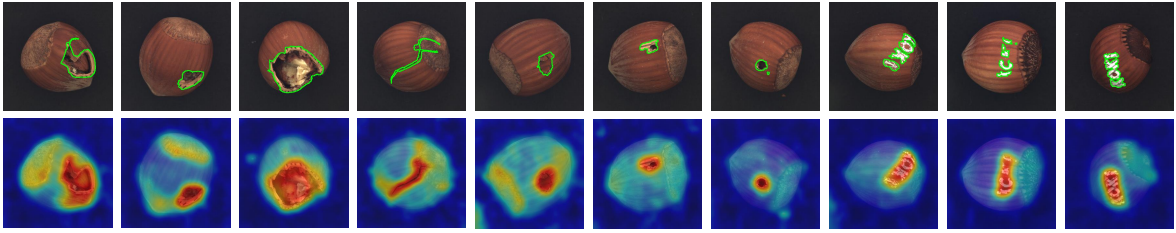}
	\caption{Visualization of segmentation results for the \textbf{hazelnut} class on \textbf{MVTecAD} under the 4-shot setting. The first row displays the input images, with green outlines indicating the ground truth regions. The second row presents the anomaly segmentation results.}
	\label{fig13}
\end{figure*}

\begin{figure*}[h]
	\centering
	\includegraphics[width=0.9\columnwidth]{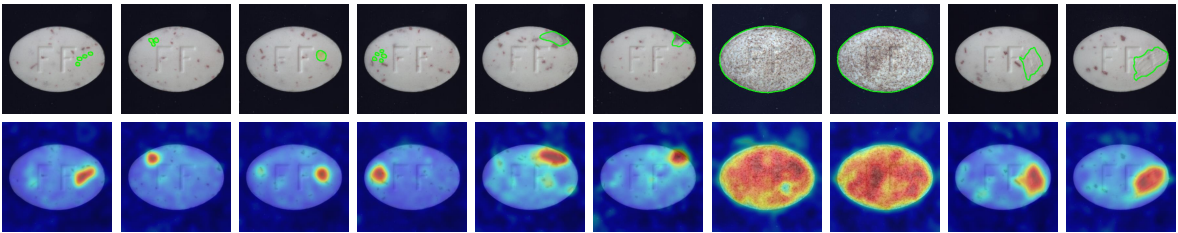}
	\caption{Visualization of segmentation results for the \textbf{pill} class on \textbf{MVTecAD} under the 4-shot setting. The first row displays the input images, with green outlines indicating the ground truth regions. The second row presents the anomaly segmentation results.}
	\label{fig14}
\end{figure*}

\begin{figure*}[h]
	\centering
	\includegraphics[width=0.9\columnwidth]{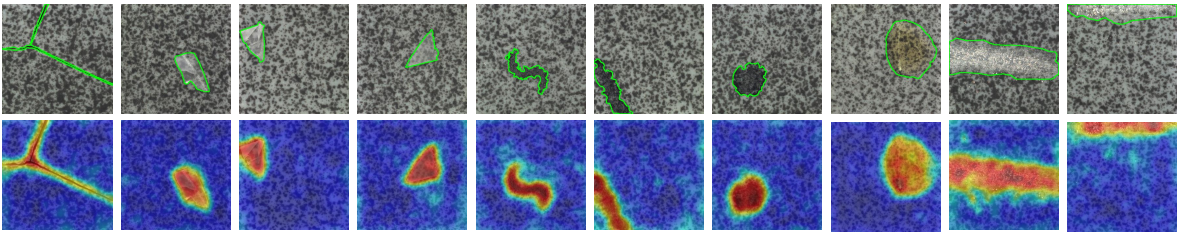}
	\caption{Visualization of segmentation results for the \textbf{tile} class on \textbf{MVTecAD} under the 4-shot setting. The first row displays the input images, with green outlines indicating the ground truth regions. The second row presents the anomaly segmentation results.}
	\label{fig15}
\end{figure*}

\begin{figure*}[h]
	\centering
	\includegraphics[width=0.9\columnwidth]{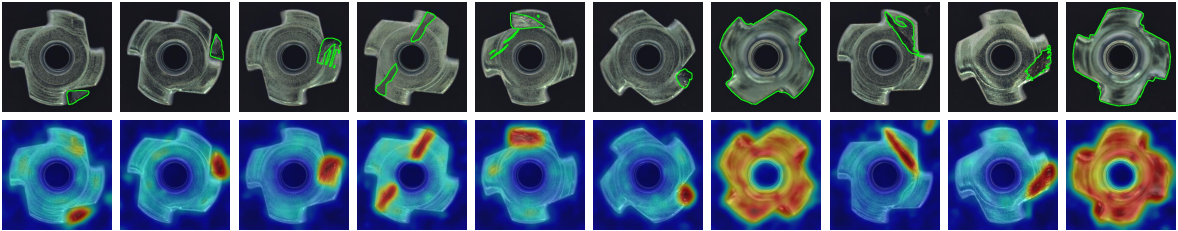}
	\caption{Visualization of segmentation results for the \textbf{metal\_nut} class on \textbf{MVTecAD} under the 4-shot setting. The first row displays the input images, with green outlines indicating the ground truth regions. The second row presents the anomaly segmentation results.}
	\label{fig16}
\end{figure*}

\begin{figure*}[h]
	\centering
	\includegraphics[width=0.9\columnwidth]{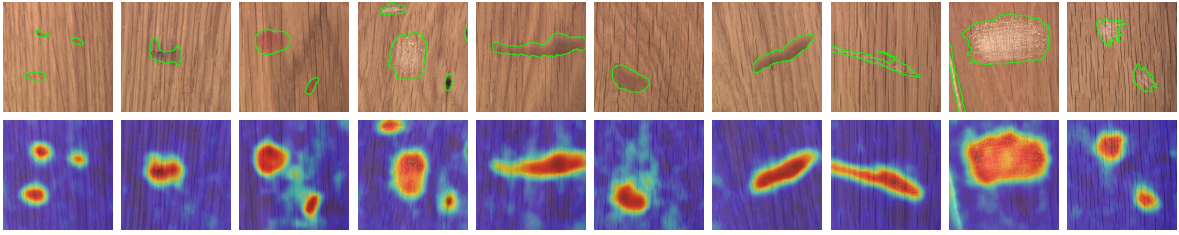}
	\caption{Visualization of segmentation results for the \textbf{wood} class on \textbf{MVTecAD} under the 4-shot setting. The first row displays the input images, with green outlines indicating the ground truth regions. The second row presents the anomaly segmentation results.}
	\label{fig17}
\end{figure*}

\begin{figure*}[h]
	\centering
	\includegraphics[width=0.9\columnwidth]{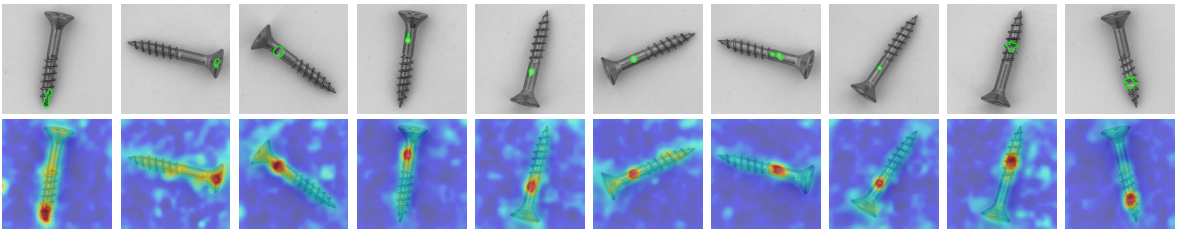}
	\caption{Visualization of segmentation results for the \textbf{screw} class on \textbf{MVTecAD} under the 4-shot setting. The first row displays the input images, with green outlines indicating the ground truth regions. The second row presents the anomaly segmentation results.}
	\label{fig18}
\end{figure*}

\begin{figure*}[h]
	\centering
	\includegraphics[width=0.9\columnwidth]{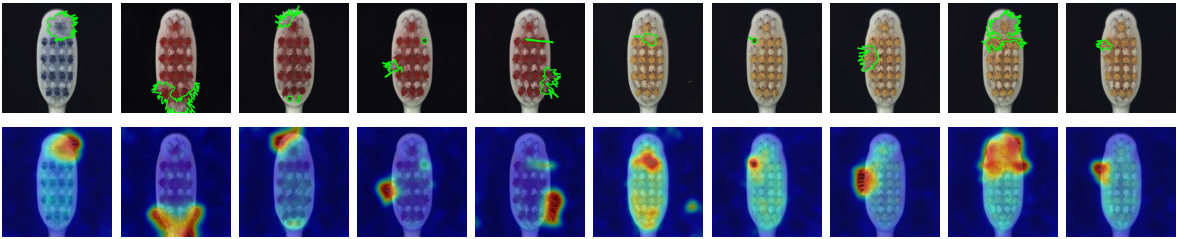}
	\caption{Visualization of segmentation results for the \textbf{toothbrush} class on \textbf{MVTecAD} under the 4-shot setting. The first row displays the input images, with green outlines indicating the ground truth regions. The second row presents the anomaly segmentation results.}
	\label{fig19}
\end{figure*}

\begin{figure*}[h]
	\centering
	\includegraphics[width=0.9\columnwidth]{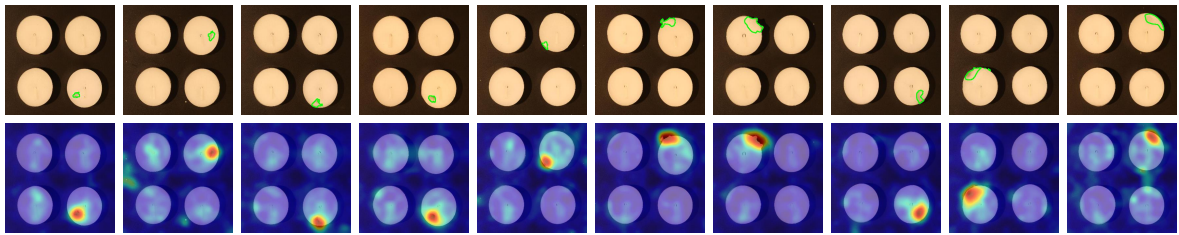}
	\caption{Visualization of segmentation results for the \textbf{candle} class on \textbf{VisA} under the 4-shot setting. The first row displays the input images, with green outlines indicating the ground truth regions. The second row presents the anomaly segmentation results.}
	\label{fig20}
\end{figure*}

\begin{figure*}[h]
	\centering
	\includegraphics[width=0.9\columnwidth]{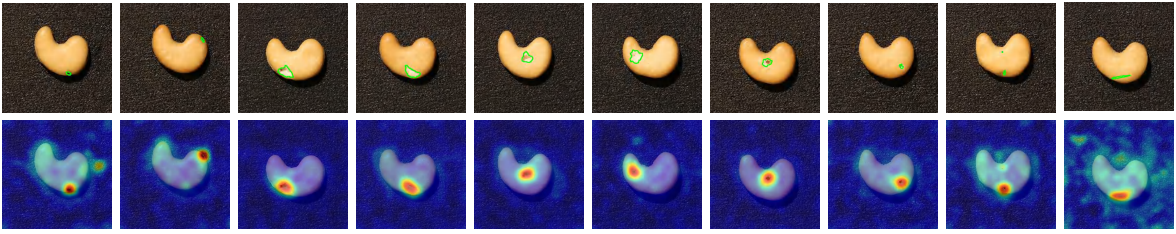}
	\caption{Visualization of segmentation results for the \textbf{cashew} class on \textbf{VisA} under the 4-shot setting. The first row displays the input images, with green outlines indicating the ground truth regions. The second row presents the anomaly segmentation results.}
	\label{fig21}
\end{figure*}

\begin{figure*}[h]
	\centering
	\includegraphics[width=0.9\columnwidth]{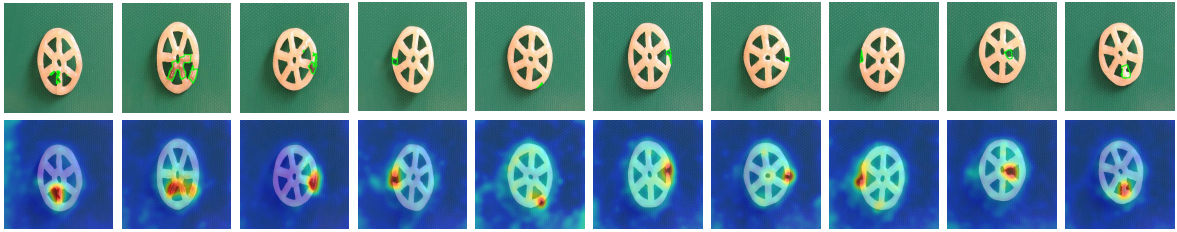}
	\caption{Visualization of segmentation results for the \textbf{fryum} class on \textbf{VisA} under the 4-shot setting. The first row displays the input images, with green outlines indicating the ground truth regions. The second row presents the anomaly segmentation results.}
	\label{fig22}
\end{figure*}

\begin{figure*}[h]
	\centering
	\includegraphics[width=0.9\columnwidth]{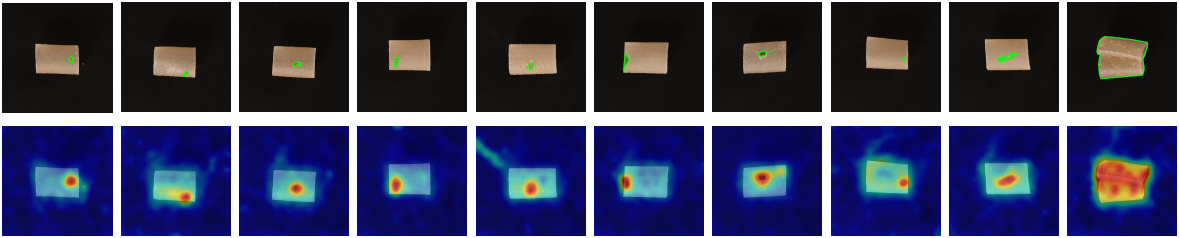}
	\caption{Visualization of segmentation results for the \textbf{pipe\_fryum} class on \textbf{VisA} under the 4-shot setting. The first row displays the input images, with green outlines indicating the ground truth regions. The second row presents the anomaly segmentation results.}
	\label{fig23}
\end{figure*}

\begin{figure*}[h]
	\centering
	\includegraphics[width=0.9\columnwidth]{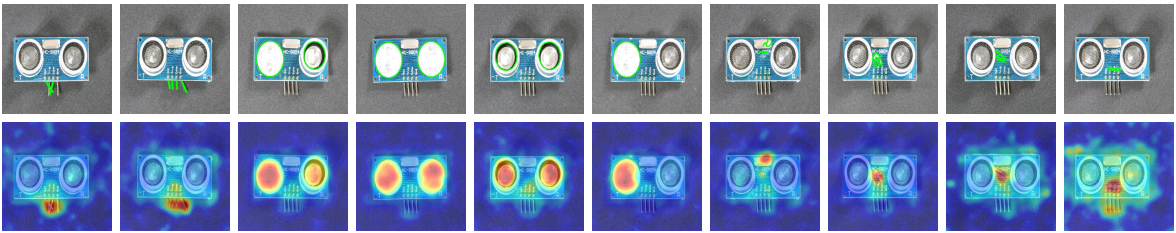}
	\caption{Visualization of segmentation results for the \textbf{PCB1} class on \textbf{VisA} under the 4-shot setting. The first row displays the input images, with green outlines indicating the ground truth regions. The second row presents the anomaly segmentation results.}
	\label{fig24}
\end{figure*}

\begin{figure*}[h]
	\centering
	\includegraphics[width=0.9\columnwidth]{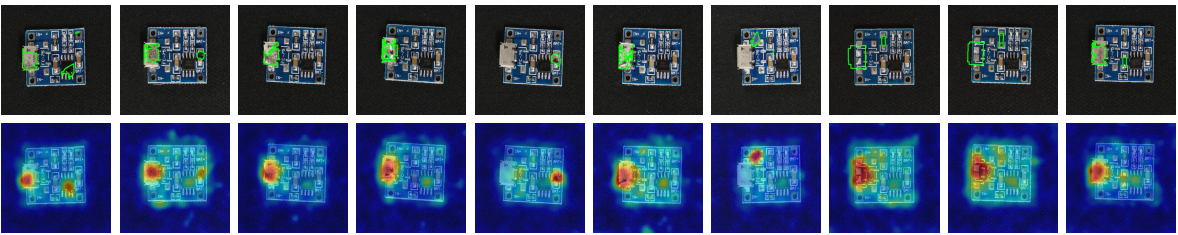}
	\caption{Visualization of segmentation results for the \textbf{PCB4} class on \textbf{VisA} under the 4-shot setting. The first row displays the input images, with green outlines indicating the ground truth regions. The second row presents the anomaly segmentation results.}
	\label{fig25}
\end{figure*}

\begin{figure*}[h]
	\centering
	\includegraphics[width=0.9\columnwidth]{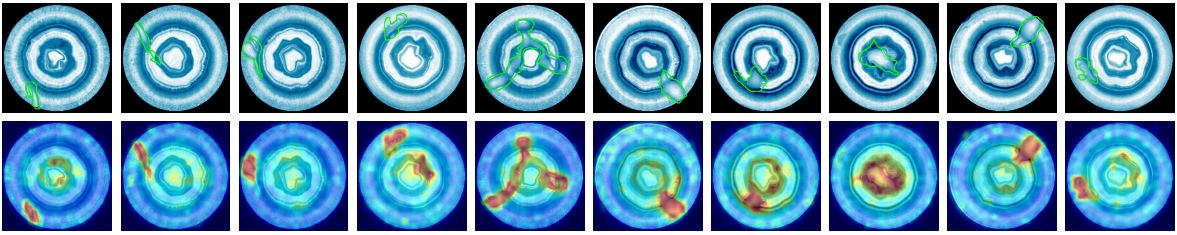}
	\caption{Visualization of segmentation results on BTAD under the 4-shot setting. The first row displays the input images, with green outlines indicating the ground truth regions. The second row presents the anomaly segmentation results..}
	\label{fig26}
\end{figure*}

\begin{figure*}[h]
	\centering
	\includegraphics[width=0.9\columnwidth]{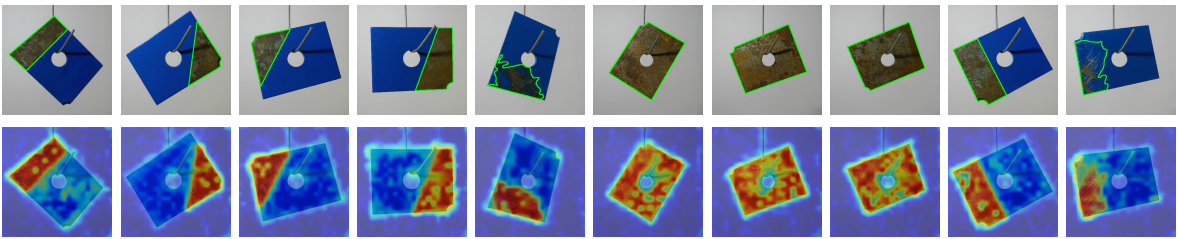}
	\caption{Visualization of segmentation results for the \textbf{metal\_plate} class on \textbf{MPDD} under the 4-shot setting. The first row displays the input images, with green outlines indicating the ground truth regions. The second row presents the anomaly segmentation results.}
	\label{fig27}
\end{figure*}

\begin{figure*}[h]
	\centering
	\includegraphics[width=0.9\columnwidth]{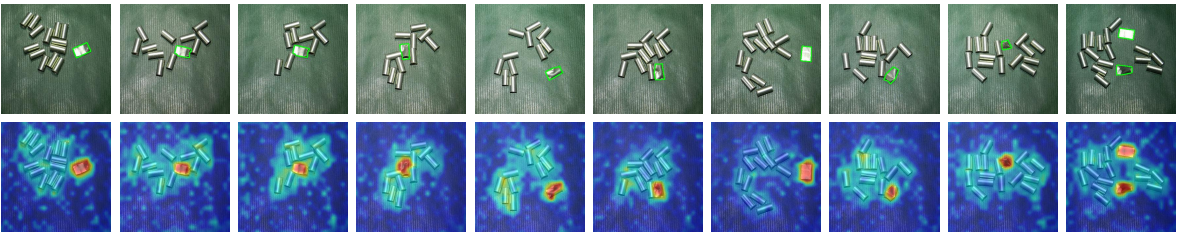}
	\caption{Visualization of segmentation results for the \textbf{tubes} class on \textbf{MPDD} under the 4-shot setting. The first row displays the input images, with green outlines indicating the ground truth regions. The second row presents the anomaly segmentation results.}
	\label{fig28}
\end{figure*}

\begin{figure*}[h]
	\centering
	\includegraphics[width=0.9\columnwidth]{./pic/bangel.pdf}
	\caption{Visualization of segmentation results for the \textbf{bangel} class on \textbf{MVTec3D} under the 4-shot setting. The first row displays the input images, with green outlines indicating the ground truth regions. The second row presents the anomaly segmentation results.}
	\label{fig29}
\end{figure*}

\begin{figure*}[h]
	\centering
	\includegraphics[width=0.9\columnwidth]{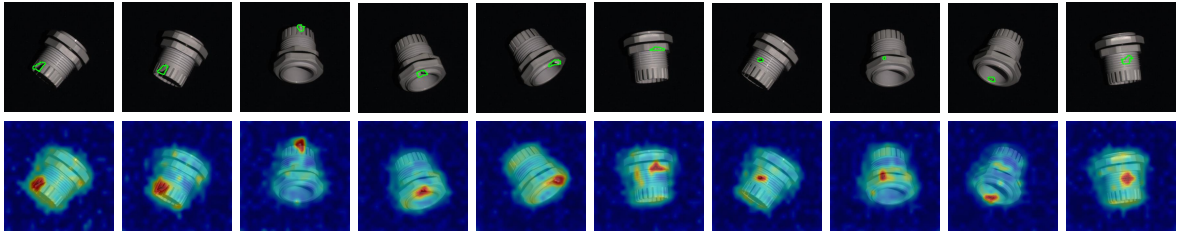}
	\caption{Visualization of segmentation results for the \textbf{cable\_gland} class on \textbf{MVTec3D} under the 4-shot setting. The first row displays the input images, with green outlines indicating the ground truth regions. The second row presents the anomaly segmentation results.}
	\label{fig30}
\end{figure*}

\begin{figure*}[h]
	\centering
	\includegraphics[width=0.9\columnwidth]{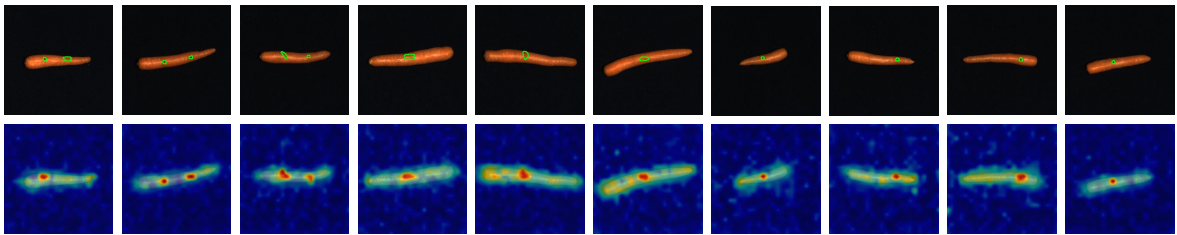}
	\caption{Visualization of segmentation results for the \textbf{carrot} class on \textbf{MVTec3D} under the 4-shot setting. The first row displays the input images, with green outlines indicating the ground truth regions. The second row presents the anomaly segmentation results.}
	\label{fig31}
\end{figure*}

\begin{figure*}[h]
	\centering
	\includegraphics[width=0.9\columnwidth]{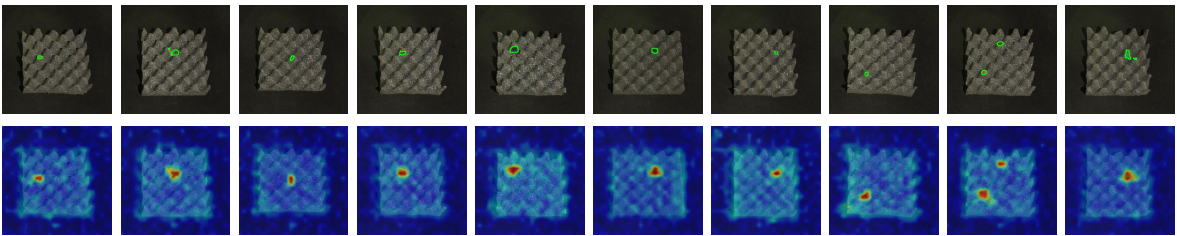}
	\caption{Visualization of segmentation results for the \textbf{foam} class on \textbf{MVTec3D} under the 4-shot setting. The first row displays the input images, with green outlines indicating the ground truth regions. The second row presents the anomaly segmentation results.}
	\label{fig32}
\end{figure*}

\begin{figure*}[h]
	\centering
	\includegraphics[width=0.9\columnwidth]{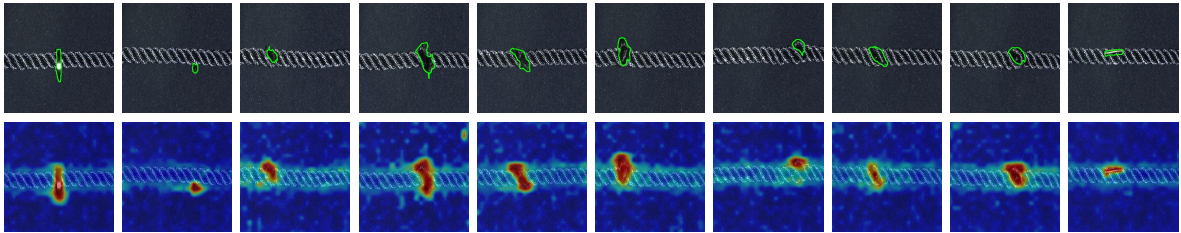}
	\caption{Visualization of segmentation results for the \textbf{rope} class on \textbf{MVTec3D} under the 4-shot setting. The first row displays the input images, with green outlines indicating the ground truth regions. The second row presents the anomaly segmentation results.}
	\label{fig33}
\end{figure*}

\begin{figure*}[h]
	\centering
	\includegraphics[width=0.9\columnwidth]{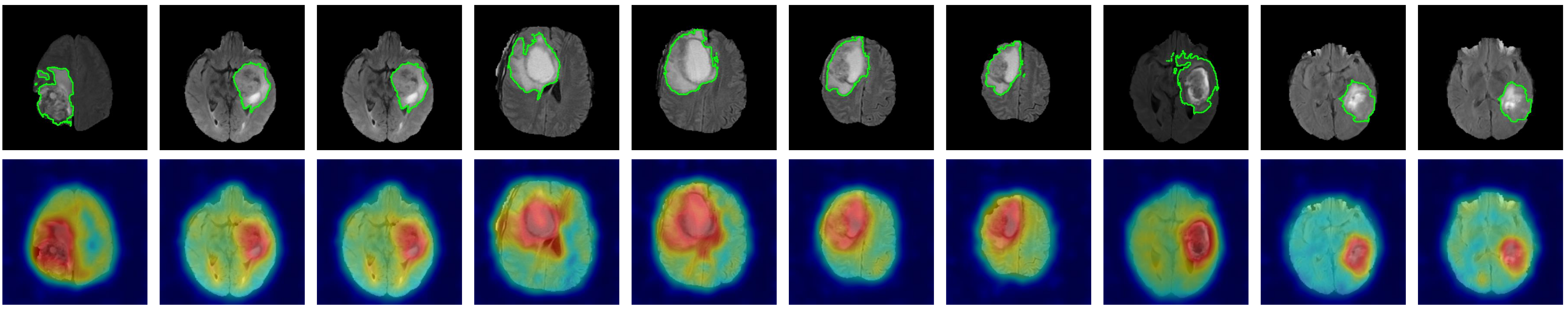}
	\caption{Visualization of segmentation results for the \textbf{brain} class on \textbf{RESC} under the 4-shot setting. The first row displays the input images, with green outlines indicating the ground truth regions. The second row presents the anomaly segmentation results.}
	\label{fig34}
\end{figure*}

\begin{figure*}[h]
	\centering
	\includegraphics[width=0.9\columnwidth]{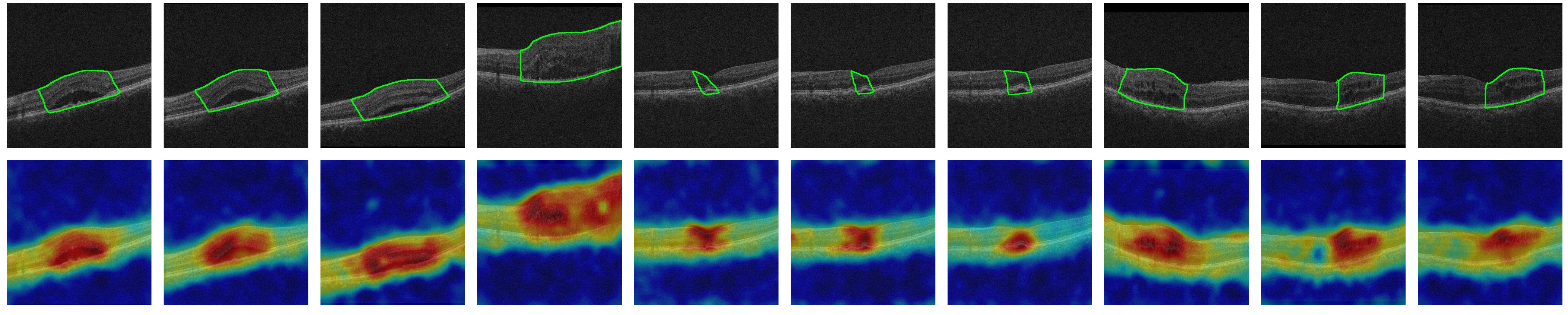}
	\caption{Visualization of segmentation results for the \textbf{retina} class on \textbf{BrasTS} under the 4-shot setting. The first row displays the input images, with green outlines indicating the ground truth regions. The second row presents the anomaly segmentation results.}
	\label{fig35}
\end{figure*}

\end{document}